%% file: iclr2027_conference.tex
\documentclass{article} 
\usepackage{iclr2027_conference,times}

\input{math_commands.tex}

\usepackage{hyperref}
\usepackage{url}

\title{\raggedright VISTA: Test-Time Compositional Alignment for Visual Autoregressive Generation}

\author{%
Hossein Shahabadi \, Niki Sepasian \, Mahdieh Soleymani Baghshah\\
Department of Computer Engineering\\
Sharif University of Technology\\
\texttt{hosseinshahabadi.me@gmail.com}
}

\iclrfinalcopy 
\usepackage{graphicx}
\usepackage{booktabs}
\usepackage{array}
\usepackage{multirow}
\usepackage[table]{xcolor}
\usepackage{float}

\definecolor{vistarow}{RGB}{232,240,250}
\definecolor{spatialpurple}{RGB}{106,61,232}
\definecolor{bindteal}{RGB}{15,110,86}
\definecolor{depthorange}{RGB}{213,94,0}

\begin{document}

\maketitle

\input{sections/0_abstract}
\input{sections/1_introduction}
\input{sections/2_preliminaries}
\input{sections/3_method}
\input{sections/4_experiments}
\input{sections/5_limitations}
\input{sections/6_conclusion}


\section*{Reproducibility Statement}
We provide the implementation used for our experiments as supplementary material. Sec.~\ref{sec:setup} specifies the backbones, benchmarks, baselines, and default configuration. Secs.~\ref{sec:capture}--\ref{sec:budget} describe the attention capture, objective interface, optimization, and scheduling mechanisms. Appendix~\ref{app:human} details the human-evaluation protocol, Appendix~\ref{app:bac} gives the complete depth-objective construction and hyperparameters, and Appendices~\ref{app:seeds} and \ref{app:overhead} report seed variance and system measurements. Appendices~\ref{app:ttsvar} and \ref{app:arguidance} document the comparison protocols, while Appendix~\ref{app:borrowed} records the definitions and provenance of the instantiated objectives. Appendices~\ref{app:fidelity}, \ref{app:budget}, and \ref{app:objvariants} provide the fidelity protocol, step-budget rule, and alternative objective instantiations. Together with the supplementary code, these details support reproduction of the reported results.

\section*{AI Use Disclosure}
In this work, we used generative AI tools to formulate mathematical
claims and assist with method implementation. We did not use generative AI tools to develop theoretical models or conceptual frameworks, provide critical ingredients for proving mathematical claims, assist in writing proofs, propose or refine hypotheses, design or provide feedback on research methodology or experiments, or interpret results. Generating synthetic datasets, translation, dataset cleaning or reformatting, and qualitative or thematic data analysis were not applicable to this work.

Additionally, we used generative AI tools to draft parts of the paper
and to aid and polish the writing. We reviewed all AI-assisted work.
AI-assisted mathematical claims were manually verified, and
AI-assisted implementation was reviewed and tested by the authors.
We take responsibility for the final content of this work, including
text, claims, code, and other artifacts produced with the aid of
generative AI.

\bibliography{iclr2027_conference}
\bibliographystyle{iclr2027_conference}

\clearpage
\appendix
\section*{Appendix}

\input{sections/8_1_relatedworks}
\input{sections/8_2_appendix}

\end{document}

%% file: math_commands.tex
\usepackage{amsmath,amsfonts,bm}

\def\eqref#1{equation~\ref{#1}}

\def\1{\bm{1}}

\DeclareMathAlphabet{\mathsfit}{\encodingdefault}{\sfdefault}{m}{sl}
\SetMathAlphabet{\mathsfit}{bold}{\encodingdefault}{\sfdefault}{bx}{n}



%% file: sections/0_abstract.tex
\begin{abstract}
Visual autoregressive (VAR) models offer fast, high-quality text-to-image generation but still fail at attribute binding and spatial relations. Existing VAR test-time methods search sampled trajectories or apply generic logit guidance but do not directly optimize compositional constraints; gradient-based diffusion methods do not transfer to stateful, discrete, multi-resolution VAR sampling. We introduce \textbf{VISTA} (\textbf{Vi}sual Autoregressive \textbf{S}emantic \textbf{T}est-time \textbf{A}lignment), the first gradient-based test-time framework for compositional alignment in next-scale autoregressive image generation. Built on Infinity, VISTA optimizes intermediate representations through the frozen transformer without parameter updates or training. It stabilizes optimization across scales and exposes a common interface into which any differentiable constraint on cross-attention can be plugged. Across two benchmarks and model scales, VISTA improves every targeted category, raising the mean by nearly $20\%$ on a 2B backbone and almost $6\%$ on an 8B backbone, especially on spatial relations. In human evaluation, VISTA wins $83\%$ of decisive compositionality comparisons by plurality vote, with no significant preference in perceived quality; independently, ImageReward scores its outputs $20\%$ higher. VISTA enables the 2B model to surpass a backbone four times its size, showing much of the compositional gap is recoverable at test time.
\end{abstract}

%% file: sections/1_introduction.tex
\section{Introduction}
\label{sec:introduction}

Latent diffusion models~\cite{rombach2022ldm} and their transformer-based successors remain the dominant approach to open-domain text-to-image synthesis. Visual autoregressive (VAR) models offer a recent alternative: VAR~\cite{tian2024var} recasts autoregressive image generation as coarse-to-fine next-scale prediction rather than raster-scan token prediction~\cite{esser2021taming}, producing increasingly resolved token maps conditioned on all coarser ones. This formulation allowed transformer-based~\cite{vaswani2017attention} autoregressive models to surpass diffusion transformers in quality, speed, and scalability. Infinity~\cite{han2024infinity} extends it with a bitwise, infinite-vocabulary tokenizer and self-correction, reaching quality competitive with leading diffusion models.

Despite this progress, high visual quality does not guarantee compositional reliability. A systematic comparison of six text-to-image systems finds that next-scale VAR models such as Infinity fail on attribute binding, spatial reasoning, and relational tasks, and that these failures persist with model scale~\cite{shahabadi2025infinitybeyond}. Visual fidelity and parameter count therefore do not imply compositional understanding. Because retraining a backbone is expensive and model-specific, a growing body of work instead intervenes at inference time.

In diffusion models, inference-time compositional alignment is already well established. Building on evidence that cross-attention encodes correspondence between image regions and text~\cite{hertz2022prompttoprompt}, Attend-and-Excite~\cite{chefer2023attendexcite} and SynGen~\cite{rassin2023syngen} optimize the diffusion latent during sampling to reshape attention toward prompt constraints. Subsequent work refines these objectives for attribute binding~\cite{meral2023conform,li2023dividebind}, concept separation~\cite{agarwal2023astar}, and spatial relations~\cite{rezaei2025psg,han2025spatial}. These methods share the premise that a frozen model's cross-attention provides a differentiable handle on compositional structure, making inference-time steering cheaper and more portable than retraining.

Test-time intervention has also reached autoregressive image generation through search and logit guidance. TTS-VAR~\cite{chen2025ttsvar} searches next-scale trajectories using coarse-scale diversity and fine-scale reward selection; GridAR~\cite{park2026progress} progressively prunes partial raster-scan generations; and ScalingAR~\cite{chen2025go} adapts test-time scaling to next-token generation. For next-scale VARs, SSG~\cite{shin2026ssg} amplifies scale-specific residual detail, while VPG~\cite{liao2026vpg} contrasts predictions from genuine and corrupted visual prefixes. These guidance methods target generic scale refinement or prefix consistency rather than explicit compositional constraints. Selection is most effective when satisfactory candidates already occur with appreciable probability, but offers limited help when the backbone fails systematically. Gradient steering can directly modify the trajectory, but cannot be applied unchanged to next-scale VAR generation. Infinity produces discrete tokens over increasing resolutions and maintains persistent state on which later scales depend. Consequently, global gradient normalization dilutes updates across increasingly large token maps until individual elements no longer affect sampling, while repeated optimization passes alter the attention cache read by subsequent scales.

We introduce \textbf{VISTA} (\textbf{Vi}sual Autoregressive \textbf{S}emantic \textbf{T}est-time \textbf{A}lignment), the first gradient-based test-time framework for compositional alignment in next-scale VAR generation. Applied to Infinity, VISTA optimizes intermediate token representations through the frozen transformer using differentiable cross-attention capture and cache-safe, scale-aware optimization. The same capture and optimization machinery serves each instantiated objective. A pluggable registry supports differentiable cross-attention constraints; we instantiate it for attribute binding, planar relations, and depth ordering.

On T2I-CompBench, VISTA improves every targeted category, including an $89\%$ relative gain on 2D spatial relations and $11$--$21\%$ gains on color, texture, and shape binding. On GenEval, VISTA raises Position from $0.250$ to $0.745$, whereas TTS-VAR reports a $7.5\%$ Position gain on Infinity-2B under its own protocol~\cite{chen2025ttsvar}. VISTA on Infinity-2B surpasses the $4\times$ larger Infinity-8B on four of six targeted categories and on their average, while applying VISTA to Infinity-8B improves that model further. Human evaluation prefers VISTA in $83\%$ of decisive compositionality comparisons and finds no significant preference in perceived quality; independently, ImageReward scores its outputs nearly $20\%$ higher. These gains trade against inference cost, which VISTA exposes through two explicit controls, while requiring no additional training, trainable parameters, reward model, or task-specific data. Appendix~\ref{app:relatedwork} discusses related work in full.

\textbf{Our contributions are threefold.}
\textbf{(i) Gradient-based alignment for next-scale VARs.} We introduce the first gradient-based test-time framework for compositional alignment in next-scale VARs, requiring no additional training, parameters, reward model, or candidate search.
\textbf{(ii) Stable optimization inside stateful generation.} We identify the failure modes that make latent optimization unstable in multi-scale VAR sampling and introduce shared capture and optimization mechanisms that address them (Secs.~\ref{sec:capture} and \ref{sec:stability}).
\textbf{(iii) A reusable objective interface.} We make differentiable cross-attention objectives developed for diffusion applicable to next-scale VARs, opening a new direction for test-time objective design. We demonstrate this interface using prior binding and planar objectives together with a new boundary-containment depth constraint.

%% file: sections/2_preliminaries.tex
\section{Preliminaries}
\label{sec:preliminaries}

\paragraph{VARs generate coarse-to-fine.}
VAR~\cite{tian2024var} represents an image as a sequence of discrete token maps $r_1,\dots,r_S$ at increasing resolutions $h_s\times w_s$. Each $r_s$ indexes a codebook $V$ learned by a VQ tokenizer, and generation factorizes autoregressively over scales:
\begin{equation}
p(r_1,\dots,r_S\mid c)
=
\prod_{s=1}^{S}
p(r_s\mid r_1,\dots,r_{s-1},c).
\end{equation}
Infinity~\cite{han2024infinity} extends VAR from class conditioning to free-form text by taking $c$ from a frozen text encoder, and replaces the finite codebook with a bitwise, infinite-vocabulary tokenizer whose labels are predicted one bit at a time. Its default $13$-scale schedule progresses from $1{\times}1$ and $2{\times}2$ grids through $4{\times}4$, $6{\times}6$, and $8{\times}8$, ultimately reaching $64{\times}64$.

\paragraph{VISTA intervenes before irreversible sampling in a stateful process.}
At scale $s$, Infinity upsamples the previously sampled and quantized codes $r_{<s}$ to resolution $h_s\times w_s$, producing a continuous representation $\mathbf{z}_s\in\mathbb{R}^{L_s\times C}$, where $L_s=h_sw_s$. A transformer processes $\mathbf{z}_s$ through self-attention over spatial positions and cross-attention to $c$, then produces logits from which $r_s$ is sampled using top-$k$/top-$p$ sampling. The resulting tokens are quantized and passed to scale $s+1$. Keys and values computed at each scale are cached and read by every later scale, so generation carries state across the entire schedule rather than restarting at each resolution. Consequently, $\mathbf{z}_s$ is a transient, per-scale representation rather than a single latent refined throughout generation as in diffusion; intervention occurs before an irreversible discrete sampling step, and every optimization forward pass can modify state needed by later scales. These properties motivate the cache-safe optimization of Secs.~\ref{sec:capture}--\ref{sec:stability}.

\paragraph{Compositionality is constraint satisfaction.}
We treat a prompt as asserting constraints over its entities: each entity's \emph{cardinality}, its bound \emph{attributes}, and its \emph{relations} to other entities. A generation fails compositionally when it violates one or more of these constraints, independently of its overall visual quality. Such failures remain a central challenge even for strong generative models: high visual fidelity does not ensure that all requested entities, attributes, and relations are jointly satisfied~\cite{huang2023t2icompbench,ghosh2023geneval,shahabadi2025infinitybeyond}.

%% file: sections/3_method.tex
\section{Method}
\label{sec:method}

\subsection{Overview}

\begin{figure}[t]
\centering
\includegraphics[width=\textwidth]{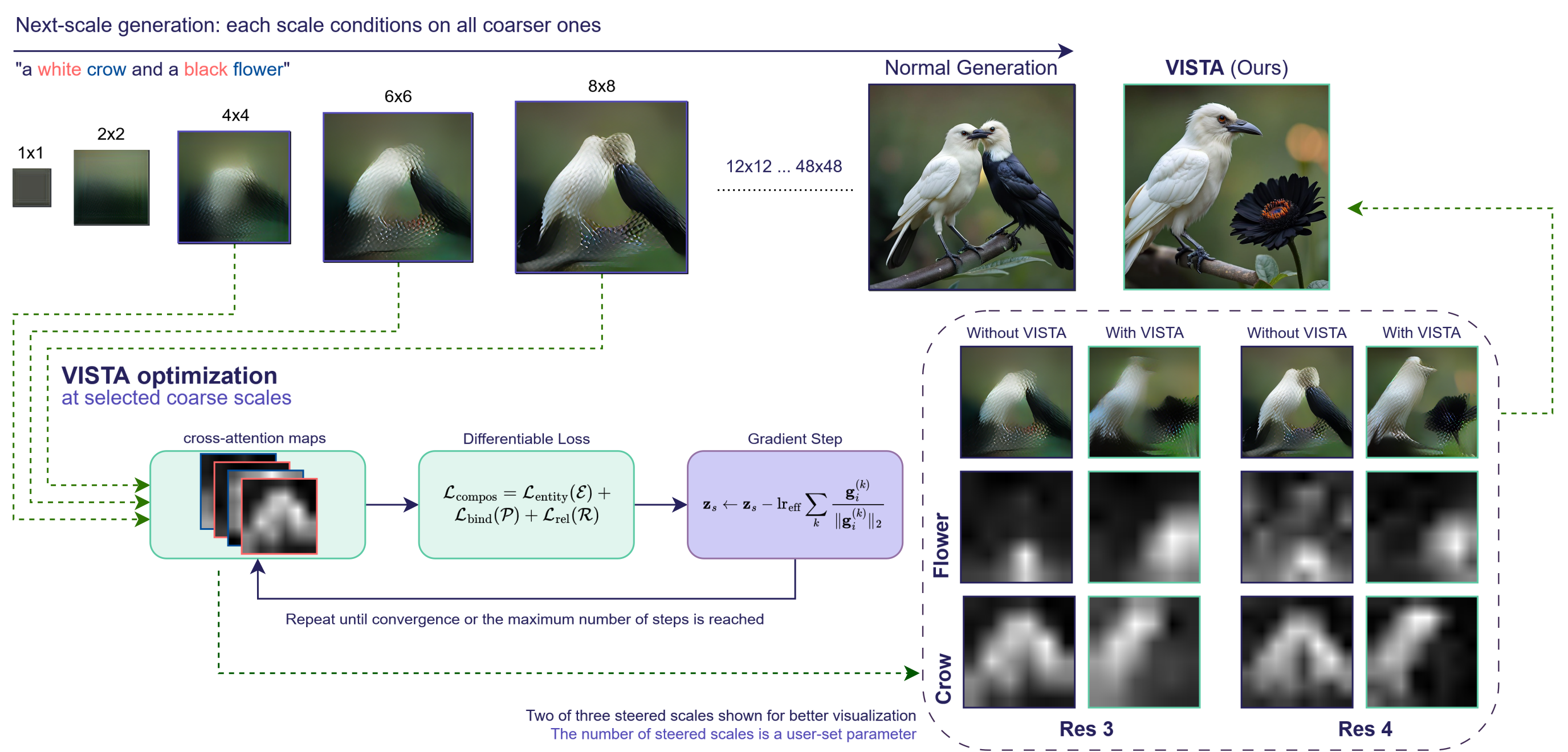}
\caption{VISTA on ``a white crow and a black flower.'' Generation proceeds coarse to fine, with every scale conditioned on all coarser ones. At the outlined scales, VISTA evaluates differentiable constraints on the model's cross-attention and updates $\mathbf{z}_s$ until all active gradients fall below their thresholds or the step budget is exhausted. \textbf{Right:} \emph{flower} and \emph{crow} attention at the $6{\times}6$ and $8{\times}8$ scales. Without steering, the maps remain collapsed and a second bird replaces the flower; with steering, they separate and remain distinct through the unsteered finer scales. $\mathcal{L}_{\mathrm{entity}}$ is shown for completeness but is not instantiated in our experiments.}
\label{fig:vista}
\end{figure}

\textbf{VISTA separates reusable optimization machinery from pluggable compositional objectives.}
Cross-attention provides the control signal: because image tokens query the text conditioning $\mathbf{c}$, its maps expose which text tokens influence each spatial region~\cite{hertz2022prompttoprompt}. At selected scales, VISTA treats Infinity's continuous input representation $\mathbf{z}_s$ as an optimization variable and steers it until these maps better satisfy constraints extracted from the prompt. The backbone and text encoder remain frozen, and no training data is required. The optimized representation $\widetilde{\mathbf{z}}_s$ replaces $\mathbf{z}_s$ before discrete sampling; the resulting token map $r_s$ then conditions every finer scale. VISTA first captures cross-attention differentiably (Sec.~\ref{sec:capture}), expresses prompt constraints as objectives over it (Sec.~\ref{sec:objectives}), and then optimizes those objectives safely inside stateful generation (Sec.~\ref{sec:stability}). Figure~\ref{fig:vista} summarizes the method.

\subsection{Differentiable Cross-Attention Capture}
\label{sec:capture}

\paragraph{Cross-attention becomes a differentiable control interface.}
Infinity normally evaluates cross-attention inside a frozen, cached sampling loop that retains no gradient path to $\mathbf{z}_s$. We reconstruct the attention probabilities during an optimization forward pass. For cross-attention layer $\ell$ and head $h$, let $\mathbf{h}_{\ell}^{(s)}(\widetilde{\mathbf{z}}_s)$ denote the current layer input induced by the representation being optimized. We compute
\begin{equation}
\mathbf{q}_{\ell h}
=
W_{\ell h}^{Q}\mathbf{h}_{\ell}^{(s)}(\widetilde{\mathbf{z}}_s),
\qquad
\mathbf{k}_{\ell h}
=
\operatorname{sg}\!\left(W_{\ell h}^{K}\mathbf{c}\right),
\qquad
\mathbf{A}_{\ell h}^{(s)}
=
\operatorname{softmax}\!\left(
\frac{\mathbf{q}_{\ell h}\mathbf{k}_{\ell h}^{\top}}{\sqrt{d_h}}
\right),
\end{equation}
where $d_h$ is the head dimension and $\operatorname{sg}(\cdot)$ denotes stop-gradient. Gradients flow through each layer's query and hidden representation to $\widetilde{\mathbf{z}}_s$, but not into the fixed text stream or model parameters. Averaging $\mathbf{A}_{\ell h}^{(s)}$ over all cross-attention layers and heads gives
$\mathbf{A}^{(s)}\in\mathbb{R}^{L_s\times L_k}$,
where $L_k$ is the text-token length. Together with the losses defined in Sec.~\ref{sec:objectives}, the captured cross-attention maps provide the signal VISTA uses to improve alignment between the text prompt and the generated image.

\subsection{Compositional Objectives}
\label{sec:objectives}

\paragraph{Prompt constraints become attention objectives.}
A prompt is compositionally satisfied only when every asserted constraint holds: each entity appears in the stated number, each attribute binds to its intended entity, and each relation holds between the correct pair. Cross-attention connects these constraints to image layout: column $t$ of $\mathbf{A}^{(s)}$ scores how strongly each spatial position attends to text token $t$. A constraint on an entity's placement can therefore be written as a differentiable function of its attention mass.

\paragraph{A common interface separates objectives from optimization.}
The parser returns explicitly counted entities $\mathcal{E}$, attribute--noun pairs $\mathcal{P}$, and ordered relation triples
$r=(X,\rho,Y)\in\mathcal{R}$.
The relation phrase $\rho$ assigns each triple to one of three disjoint sets:
$\mathcal{R}_{\mathrm{2D}}$ for image-plane relations,
$\mathcal{R}_{\mathrm{dep}}$ for viewing-axis relations, and
$\mathcal{R}_{\mathrm{ns}}$ for non-spatial relations. The general objective is
\begin{equation}
\mathcal{L}_\mathrm{compos} = \mathcal{L}_\mathrm{entity}(\mathcal{E}) + \mathcal{L}_\mathrm{bind}(\mathcal{P}) + \mathcal{L}_\mathrm{rel}(\mathcal{R}),
\qquad
\mathcal{L}_\mathrm{rel} = \underbrace{\mathcal{L}_\mathrm{2D} + \mathcal{L}_\mathrm{depth}}_{\mathcal{L}_\mathrm{spatial}} + \mathcal{L}_\mathrm{ns},
\label{eq:compos_general}
\end{equation}
An empty set contributes zero. Our reported configuration instantiates
$\mathcal{L}_{\mathrm{bind}}$,
$\mathcal{L}_{\mathrm{2D}}$, and
$\mathcal{L}_{\mathrm{depth}}$;
$\mathcal{L}_{\mathrm{entity}}$ and
$\mathcal{L}_{\mathrm{ns}}$ remain inactive.

For a text-token group $T$, define its attention map at scale $s$ as
\begin{equation}
\mathbf{m}_{T}^{(s)}
=
\frac{1}{|T|}
\sum_{t\in T}
\mathbf{A}_{:,t}^{(s)}
\in\mathbb{R}^{L_s}.
\end{equation}
We drop the scale superscript when evaluating one scale and write $\mathbf{m}_X$ for entity $X$. The interface accepts any differentiable scalar objective over these maps.

\paragraph{Existing diffusion objectives transfer through the interface.}
The interface accepts any differentiable scalar objective over the captured attention maps without changing the capture or optimization machinery. Throughout this work, we instantiate it with objectives for attribute binding, planar relations, and depth ordering. For binding, $\mathcal{L}_{\mathrm{bind}}$ adapts CONFORM's multi-positive contrastive objective~\cite{meral2023conform}: an attribute and its noun form a positive pair, while other pairs provide negatives, pulling each attribute toward its intended entity and away from competing objects. For planar relations, $\mathcal{L}_{\mathrm{2D}}$ adopts PSG's probability-of-superiority objective~\cite{rezaei2025psg}, which interprets two attention maps as spatial distributions and drives their ordering along the asserted axis. We introduce the occlusion-based depth objective below. Appendix~\ref{app:borrowed} gives the complete definitions and provenance of the binding and planar terms, while Appendix~\ref{app:objvariants} evaluates alternative objectives under the same framework.

\paragraph{Depth ordering is induced through occlusion geometry.}
Cross-attention has no explicit depth coordinate, but front--back order leaves an asymmetric footprint in the image plane. For ``$B$ behind $A$,'' our depth objective combines a separation term that prevents the two attention supports from collapsing and a boundary-containment term that encourages the portion of $B$ facing $A$ to pass through $A$'s support:
\begin{equation}
\mathcal{L}_{\mathrm{depth}}
=
\lambda_{\mathrm{sep}}\mathcal{L}_{\mathrm{sep}}
+
\lambda_{\mathrm{bc}}\mathcal{L}_{\mathrm{bc}} .
\end{equation}
This does not estimate metric depth; it promotes the overlap and foreground--background asymmetry through which the requested order becomes visible. Appendix~\ref{app:bac} gives the complete construction, motivation, and hyperparameters.

This objective is one instantiation proposed alongside VISTA, not a requirement of the framework. Appendix~\ref{app:objvariants} holds the capture and optimization machinery fixed and compares five depth objectives; every alternative enters through the same interface, and boundary containment performs best in that diagnostic comparison.

\subsection{Stable Optimization in a Stateful Generation Process}
\label{sec:stability}

Given the active objectives above, VISTA must update $\mathbf{z}_s$ without corrupting Infinity's persistent state, suppressing useful gradients, or spending equal computation on every scale.

\paragraph{Cache isolation preserves the sampling trajectory.}
Each auxiliary optimization forward writes to the self-attention cache used by later scales. VISTA therefore snapshots the cache before optimization, restores it before every auxiliary forward, and restores it once more before ordinary sampling resumes. The backbone remains frozen, and the final sampling forward sees the same pre-scale state it would have received without optimization.

\paragraph{Per-token directions balance heterogeneous objectives.}
\label{sec:gradnorm}
\label{sec:combine}
Let $t$ index optimization steps, $i\in\{1,\ldots,L_s\}$ spatial tokens, and $k$ active objectives. For objective $\mathcal{L}_k$, define
\begin{equation}
\mathbf{g}_{i}^{(k,t)}
=
\nabla_{\widetilde{\mathbf{z}}_{s,i}^{(t)}}\mathcal{L}_k,
\qquad
\overline{g}^{(k,t)}
=
\frac{1}{L_s}
\sum_{i=1}^{L_s}
\left\lVert
\mathbf{g}_{i}^{(k,t)}
\right\rVert_2 .
\end{equation}
Let
$
\mathcal{K}_t
=
\{k:\overline{g}^{(k,t)}\geq\tau_k\},
$
where $\tau_k$ is the activity threshold for objective $k$. Starting from
$\widetilde{\mathbf{z}}_s^{(0)}=\mathbf{z}_s$,
VISTA combines normalized gradient directions rather than incomparable loss values:
\begin{equation}
\mathbf{d}_i^{(t)}
=
\sum_{k\in\mathcal{K}_t}
\frac{\mathbf{g}_{i}^{(k,t)}}
{\left\lVert
\mathbf{g}_{i}^{(k,t)}
\right\rVert_2+\epsilon},
\qquad
\widetilde{\mathbf{z}}_{s,i}^{(t+1)}
=
\widetilde{\mathbf{z}}_{s,i}^{(t)}
-
\eta_s\mathbf{d}_i^{(t)} ,
\end{equation}
where $\epsilon>0$ is a small constant. All objectives share one forward graph and contribute to one joint update.

The scale-aware step size is
\begin{equation}
\eta_s
=
\mathrm{lr}_{\mathrm{base}}\,
\operatorname{RMS}(\mathbf{z}_s)\sqrt{C},
\end{equation}
where $C$ is the feature dimension and RMS is computed over the $L_sC$ entries of $\mathbf{z}_s$. Per-token normalization prevents updates from vanishing as the grid grows, while RMS scaling adapts one base step size to different scales and prompts.

\paragraph{Compute follows violation at structurally important scales.}
\label{sec:budget}
The initial ratios
$\overline{g}^{(k,0)}/\tau_k$
determine an effective ceiling
$1\leq N_{\mathrm{eff}}\leq N_{\max}$:
nearly satisfied scales receive few steps, while strongly violated ones receive the full budget. Optimization terminates earlier whenever $\mathcal{K}_t$ becomes empty. Appendix~\ref{app:budget} gives the exact scheduler, and Sec.~\ref{sec:experiments} reports the backbone-specific thresholds used.

VISTA additionally restricts intervention to scales where global structure is still forming. Perturbation analyses show that changing coarse VAR scales alters scene structure, whereas changing fine scales mainly affects local detail~\cite{tang2026varscaling}. We therefore begin at $4{\times}4$, exclude the geometrically degenerate $1{\times}1$ and $2{\times}2$ grids, and steer the earliest three non-degenerate scales by default. Sec.~\ref{sec:scalebudget} shows that gains saturate beyond this point.

%% file: sections/4_experiments.tex
\section{Experiments}
\label{sec:experiments}

\subsection{Setup}
\label{sec:setup}

\paragraph{Evaluation spans two Infinity model sizes and two complementary benchmarks.}
We evaluate VISTA on Infinity-2B and Infinity-8B~\cite{han2024infinity}, using each model's default 13-scale schedule, and report results on T2I-CompBench~\cite{huang2023t2icompbench} and GenEval~\cite{ghosh2023geneval}. VISTA and its corresponding Infinity baselines are averaged over four seeds (Appendix~\ref{app:seeds}); BoN and TTS-VAR are single runs under our setup, and prior-system rows are reported from~\cite{shahabadi2025infinitybeyond}. Every Infinity row, baseline and VISTA alike, uses ScaleKV~\cite{li2025scalekv} cache compression. We refer to these configurations as \textbf{Infinity-2B} and \textbf{Infinity-8B} throughout and identify uncompressed results explicitly.

\paragraph{The setup isolates VISTA's contribution.}
Because VISTA operates on top of ScaleKV, the compressed backbone is its direct reference. Appendix~\ref{app:seeds} also reports the uncompressed Infinity-2B baseline: no category changes by more than $0.008$, and the targeted average is essentially unchanged ($0.481$ vs.\ $0.480$), so compression does not explain VISTA's gains. Unless stated otherwise, VISTA activates
$\mathcal{L}_{\mathrm{bind}}+
\mathcal{L}_{\mathrm{2D}}+
\mathcal{L}_{\mathrm{depth}}$
and steers the $4{\times}4$, $6{\times}6$, and $8{\times}8$ scales with
$N_{\max}=5$. The base step size $\mathrm{lr}_{\mathrm{base}}$ and gradient threshold $\tau_k$ are set per backbone; Appendix~\ref{app:overhead} reports their values and selection details. No numeracy or non-spatial objective is active. We therefore report those categories unchanged from the corresponding baseline and exclude them from the targeted average.

\subsection{Main Results}
\label{sec:main}

Table~\ref{tab:main} reports T2I-CompBench results for both backbones and situates them against four diffusion and diffusion-transformer systems from prior benchmarking work~\cite{shahabadi2025infinitybeyond}.

\begin{table*}[t]
\centering
\caption{T2I-CompBench. VISTA and its Infinity baselines use four seeds; BoN and TTS-VAR are single runs. Prior-system rows are from~\protect\cite{shahabadi2025infinitybeyond}. Every Infinity row uses ScaleKV (Sec.~\ref{sec:setup}). VISTA steers three scales; TTS-VAR uses the authors' implementation at $N{=}1$ and $N{=}2$, and BoN uses $N{=}8$. Numeracy and Non-Spatial (shaded) are untargeted and unchanged by construction. Avg.$_6$ covers the six targeted categories and Avg.$_8$ all eight. \textbf{Bold}: best within each backbone group; \underline{underlined}: best overall.}
\label{tab:main}
\small
\setlength{\tabcolsep}{3.5pt}
\renewcommand{\arraystretch}{0.95}
\begin{tabular}{lcccccc>{\columncolor[gray]{0.92}}c>{\columncolor[gray]{0.92}}ccc}
\toprule
Model & Color & Texture & Shape & 2D & 3D & Cmplx & Num. & Non-S. & Avg.$_6$ & Avg.$_8$ \\
\midrule
SDXL            & 0.593 & 0.519 & 0.466 & 0.215 & 0.341 & 0.319 & 0.504 & 0.311 & 0.409 & 0.409 \\
PixArt-$\alpha$ & 0.407 & 0.444 & 0.367 & 0.202 & 0.350 & 0.324 & 0.506 & 0.308 & 0.349 & 0.363 \\
Flux-Dev        & 0.746 & 0.644 & 0.482 & 0.273 & 0.393 & 0.363 & 0.613 & 0.309 & 0.483 & 0.478 \\
Flux-Schnell    & 0.725 & 0.683 & 0.559 & 0.271 & 0.373 & 0.364 & 0.604 & 0.312 & 0.496 & 0.486 \\
\midrule
Infinity-2B               & 0.749 & 0.632 & 0.475 & 0.234 & 0.405 & 0.384 & 0.572 & 0.310 & 0.480 & 0.470 \\
\quad + BoN ($N{=}8$)     & 0.739 & 0.614 & 0.456 & 0.233 & 0.398 & 0.383 & 0.565 & 0.309 & 0.471 & 0.462 \\
\quad + TTS-VAR ($N{=}1$) & 0.773 & 0.690 & 0.543 & 0.269 & 0.424 & 0.391 & 0.596 & 0.311 & 0.515 & 0.500 \\
\quad + TTS-VAR ($N{=}2$) & 0.779 & 0.679 & 0.547 & 0.265 & 0.417 & 0.391 & 0.615 & 0.311 & 0.513 & 0.501 \\
\rowcolor{vistarow}
\quad + VISTA
& \textbf{0.832} & \textbf{0.750} & \textbf{0.574}
& \underline{\textbf{0.442}} & \underline{\textbf{0.452}}
& \textbf{0.398}
& \cellcolor[gray]{0.92}0.572
& \cellcolor[gray]{0.92}0.310
& \textbf{0.575} & \textbf{0.541} \\
\midrule
Infinity-8B & 0.827 & 0.753 & 0.604 & 0.365 & 0.414 & 0.397 & 0.612 & 0.316 & 0.560 & 0.536 \\
\rowcolor{vistarow}
\quad + VISTA
& \underline{0.858} & \underline{0.798} & \underline{0.658}
& 0.415 & 0.424 & \underline{0.403}
& \cellcolor[gray]{0.92}0.612
& \cellcolor[gray]{0.92}0.316
& \underline{0.593} & \underline{0.560} \\
\bottomrule
\end{tabular}
\end{table*}

\paragraph{VISTA improves every targeted category on both model sizes.}
On Infinity-2B, the largest gain is on 2D Spatial ($+88.9\%$), followed by improvements of $11.1$--$20.8\%$ on Color, Texture, and Shape, $11.6\%$ on 3D Spatial, and $3.6\%$ on Complex. Complex improves least because its prompts combine several constraints and no single objective targets their interaction. Across the six targeted categories, VISTA raises the mean by $19.8\%$ on 2B and $5.9\%$ on 8B. The smaller relative gain on 8B reflects its stronger starting point and reduced headroom. The targeted-average standard deviation is $0.002$ on both model sizes, compared with gains of $0.095$ and $0.033$, respectively (Appendix~\ref{app:seeds}).

\paragraph{Per-category results isolate the three instantiated objectives.}
VISTA activates a term only when the prompt contains its corresponding constraint. Color, Texture, and Shape therefore isolate
$\mathcal{L}_{\mathrm{bind}}$,
2D Spatial isolates
$\mathcal{L}_{\mathrm{2D}}$,
and 3D Spatial isolates
$\mathcal{L}_{\mathrm{depth}}$.
These columns provide objective-specific evidence while preserving the complete-system evaluation on Complex prompts.

\paragraph{Test-time alignment lets 2B compete with a $4\times$ larger backbone.}
VISTA on Infinity-2B surpasses unmodified Infinity-8B on four of the six targeted categories --- Color, 2D Spatial, 3D Spatial, and Complex --- and on their average ($0.575$ vs.\ $0.560$). It trails only on Texture and Shape. This suggests that some of the observed compositional gap between the two model sizes can be recovered by steering the smaller model.

\subsection{Qualitative Results}
\label{sec:qualitative}

\begin{figure*}[t]
\centering
\setlength{\tabcolsep}{1pt}
\renewcommand{\arraystretch}{0.4}
\begin{tabular}{
@{}c@{\hspace{2pt}}
*{5}{>{\centering\arraybackslash}p{0.176\textwidth}}
@{}
}
&
{\fontsize{7pt}{7.2pt}\selectfont
``a bicycle \textcolor{spatialpurple}{to the right of} a fire hydrant''
\par}
&
{\fontsize{7pt}{7.2pt}\selectfont
``a cat \textcolor{spatialpurple}{above} a dog on a staircase''
\par}
&
{\fontsize{7pt}{7.2pt}\selectfont
``a box \textcolor{depthorange}{in front of} a chair''
\par}
&
{\fontsize{7pt}{7.2pt}\selectfont
``a \textcolor{bindteal}{red banana} and a
\textcolor{bindteal}{green strawberry}''
\par}
&
{\fontsize{6.7pt}{7.2pt}\selectfont
``an elderly man wearing a \textcolor{bindteal}{striped scarf}
\textcolor{spatialpurple}{sitting to the left of} a young woman
in a \textcolor{bindteal}{green dress} ...''
\par}
\\[1pt]
\rotatebox{90}{\scriptsize\textbf{Base}}
&
\includegraphics[width=\linewidth]
{Assets/normal/a\_bicycle\_to\_the\_right\_of\_a\_fire\_hydrant.png}
&
\includegraphics[width=\linewidth]
{Assets/normal/a\_cat\_above\_a\_dog\_on\_a\_staircase.png}
&
\includegraphics[width=\linewidth]
{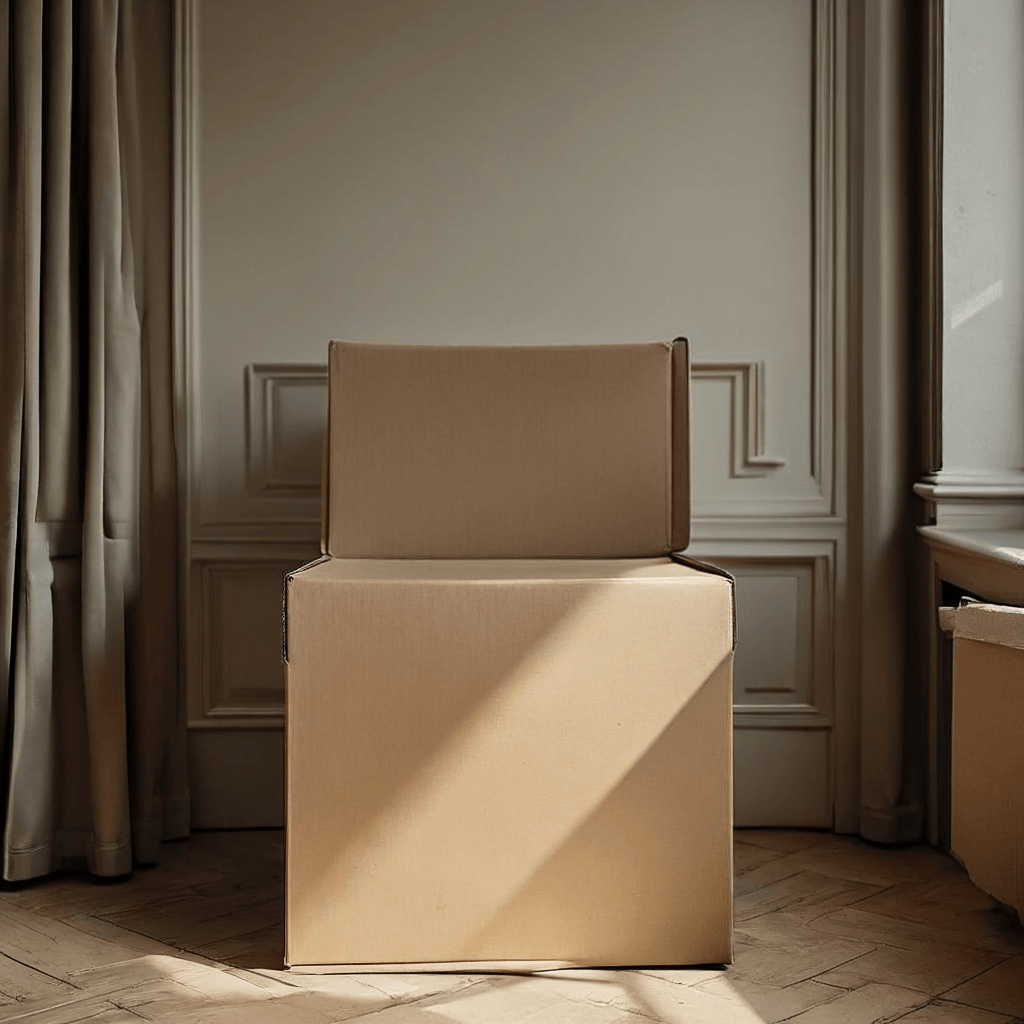}
&
\includegraphics[width=\linewidth]
{Assets/normal/a\_red\_banana\_and\_a\_green\_strawberry.png}
&
\includegraphics[width=\linewidth]
{Assets/normal/an\_elderly\_man\_wearing\_a\_striped\_scarf\_sitting\_to\_the\_left\_of\_a\_young\_woman\_in\_a\_green\_dress\_at\_a\_wooden\_table\_with\_two\_steaming\_cups\_of\_tea.png}
\\[1pt]
\rotatebox{90}{\scriptsize\textbf{VISTA (Ours)}}
&
\includegraphics[width=\linewidth]
{Assets/vista/a\_bicycle\_to\_the\_right\_of\_a\_fire\_hydrant.png}
&
\includegraphics[width=\linewidth]
{Assets/vista/a\_cat\_above\_a\_dog\_on\_a\_staircase.png}
&
\includegraphics[width=\linewidth]
{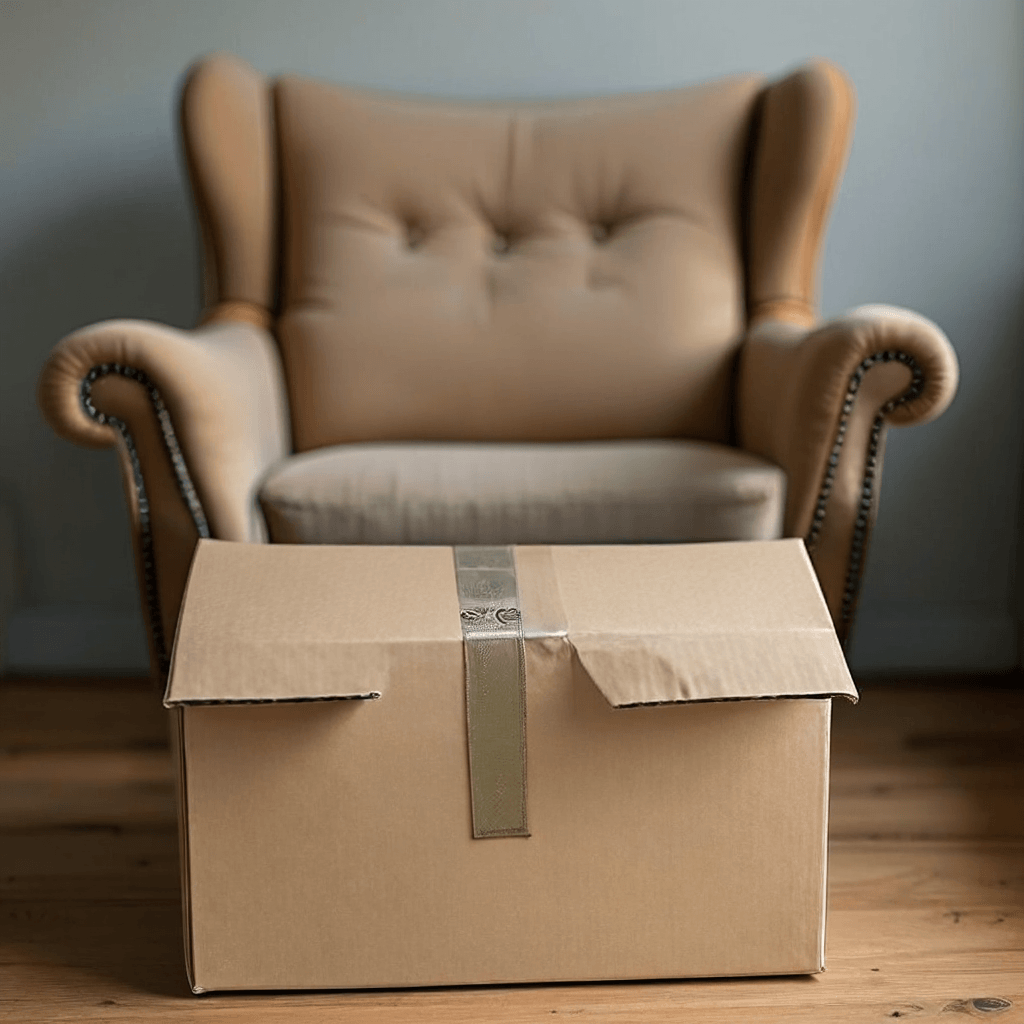}
&
\includegraphics[width=\linewidth]
{Assets/vista/a\_red\_banana\_and\_a\_green\_strawberry.png}
&
\includegraphics[width=\linewidth]
{Assets/vista/an\_elderly\_man\_wearing\_a\_striped\_scarf\_sitting\_to\_the\_left\_of\_a\_young\_woman\_in\_a\_green\_dress\_at\_a\_wooden\_table\_with\_two\_steaming\_cups\_of\_tea.png}
\\
\end{tabular}
\caption{Matched Base and VISTA generations across planar relations, depth ordering, attribute binding, and their combination. \textcolor{spatialpurple}{Purple} marks planar relations; \textcolor{depthorange}{orange} marks depth relations; \textcolor{bindteal}{teal} marks attribute--noun bindings.}
\label{fig:qualitative_grid}
\end{figure*}

\paragraph{VISTA improves isolated and joint compositional constraints.}
Figure~\ref{fig:qualitative_grid} compares Base and VISTA under matched prompts and random seeds. The first two examples isolate planar relations, the third isolates depth ordering, the fourth isolates attribute binding, and the fifth combines planar and binding constraints. In the box--chair example, Base omits the chair, whereas VISTA renders both objects with the box in front of the chair. In the combined example, Base places the man on the wrong side of the woman and leaks the striped and green attributes across subjects; VISTA corrects both constraints. The cat--dog example also shows less subject merging under VISTA, a beneficial side effect of better-separated attention maps.

\paragraph{Coarse corrections persist through unsteered scales.}
Figure~\ref{fig:vista} shows the mechanism behind these changes. The \emph{flower} and \emph{crow} maps are nearly coincident under Base, separate at the steered scales, and remain distinct through the unsteered finer scales. Correcting the coarse layout is sufficient because later scales inherit and refine it rather than reconstructing it. Appendix~\ref{app:qualitative} provides additional depth-ordering examples.

\subsection{GenEval}
\label{sec:geneval}

We evaluate the same configurations on GenEval to test whether the gains transfer beyond T2I-CompBench.

\begin{table}[t]
\centering
\caption{GenEval over four seeds. All rows use ScaleKV, and VISTA uses the configuration of Table~\ref{tab:main}. Shaded categories are untargeted and unchanged by construction. \textbf{Bold}: best within each backbone group; \underline{underlined}: best overall.}
\label{tab:geneval}
\small
\setlength{\tabcolsep}{3.5pt}
\begin{tabular}{lccc>{\columncolor[gray]{0.92}}c>{\columncolor[gray]{0.92}}c>{\columncolor[gray]{0.92}}cc}
\toprule
Model & Color & Attr. & Pos. & Single & Two & Count & Overall \\
\midrule
Infinity-2B & 0.827 & 0.560 & 0.250 & 0.994 & 0.808 & 0.591 & 0.672 \\
\rowcolor{vistarow}
\quad + VISTA
& 0.809 & \textbf{0.645} & \textbf{0.745}
& \cellcolor[gray]{0.92}0.994
& \cellcolor[gray]{0.92}0.808
& \cellcolor[gray]{0.92}0.591
& \textbf{0.765} \\
\midrule
Infinity-8B & 0.886 & 0.765 & 0.578 & 1.000 & 0.937 & 0.778 & 0.824 \\
\rowcolor{vistarow}
\quad + VISTA
& \underline{0.907} & \underline{0.788} & \underline{0.825}
& \cellcolor[gray]{0.92}1.000
& \cellcolor[gray]{0.92}0.937
& \cellcolor[gray]{0.92}0.778
& \underline{0.873} \\
\bottomrule
\end{tabular}
\end{table}

\paragraph{The gains transfer across benchmarks and model scales.}
Position improves by $198\%$ on 2B ($0.250\rightarrow0.745$) and $42.7\%$ on 8B, raising the overall score by $13.8\%$ and $5.9\%$, respectively. VISTA on 2B surpasses unmodified 8B on Position ($0.745$ vs.\ $0.578$), while attribute binding improves on both backbones. Color decreases slightly on 2B ($0.827\rightarrow0.809$), an effect absent on 8B and T2I-CompBench; GenEval's Color category measures single-object attribution rather than the multi-object binding targeted by $\mathcal{L}_{\mathrm{bind}}$. SSG reports a $+0.007$ GenEval Overall change on Infinity-2B, while VPG reports $+0.01$ Overall and $+0.02$ Position~\cite{shin2026ssg,liao2026vpg}. Because their baselines and prompt protocols differ (VPG uses rewritten prompts, whereas ours remain unchanged), Appendix~\ref{app:arguidance} compares their within-paper changes separately rather than inserting unmatched rows here.

\paragraph{Steering helps when sampling rarely contains a correct candidate.}
Table~\ref{tab:main} compares VISTA with BoN and TTS-VAR~\cite{chen2025ttsvar}. Following TTS-VAR, BoN generates multiple complete trajectories and selects the final image with the highest ImageReward score~\cite{xu2023imagereward}. This is effective when a satisfactory generation already appears with appreciable probability, but increasing the sample count offers limited help when the backbone repeatedly makes the same compositional error. The distinction is clearest for spatial relations: neither BoN nor increasing TTS-VAR from $N{=}1$ to $N{=}2$ provides consistent improvement, whereas VISTA steers the trajectory toward the requested arrangement. Sampling performs better on Numeracy, where successes vary across trajectories and VISTA activates no corresponding objective.

\paragraph{Candidate search introduces a different resource trade-off.}
TTS-VAR maintains multiple trajectories under a descending batch schedule, using DINOv2 clustering at coarse scales and ImageReward-based resampling later. Its computation and memory grow with search width and require intermediate decoding through additional pretrained models. VISTA instead optimizes one trajectory from internal cross-attention, requiring neither a candidate batch nor an external verifier. Its overhead depends on the number of steered scales and optimization steps rather than the number of sampled trajectories.

\subsection{Quality, Fidelity, and Cost}
\label{sec:scalebudget}

Table~\ref{tab:budget} varies the number of steered scales along three axes: compositional accuracy, image fidelity, and generation time.

\begin{table}[t]
\centering
\caption{Quality--cost trade-off as the number of steered scales increases coarse-to-fine from $4{\times}4$. The fidelity columns report ImageReward (prompt-conditioned) and LAION aesthetic v2 (unconditional) on $600$ paired prompts; Appendix~\ref{app:fidelity} reports additional scorers. The shaded row is our default.}
\label{tab:budget}
\small
\setlength{\tabcolsep}{5pt}
\begin{tabular}{lcc|cc|c}
\toprule
Steered scales & Targ.\ 2B $\uparrow$ & Targ.\ 8B $\uparrow$ & ImageReward 2B $\uparrow$ & Aesthetic 2B $\uparrow$ & Time 2B $\downarrow$ \\
\midrule
None & 0.480 & 0.560 & 1.025 & 5.734 & 3.34\,s \\
\midrule
1 & 0.546 & 0.579 & 1.234 & 5.683 & 3.99\,s \\
2 & 0.567 & 0.587 & 1.234 & 5.615 & 4.74\,s \\
\rowcolor{vistarow}
3 & 0.575 & 0.593 & 1.230 & 5.609 & 5.46\,s \\
4 & 0.579 & 0.586 & 1.204 & 5.608 & 6.28\,s \\
\bottomrule
\end{tabular}
\end{table}

\begin{table}[t]
\centering
\caption{Human evaluation on $200$ paired generations. Counts are plurality-aggregated outcomes, one per image pair; Win$_{\mathrm{dec.}}$ is VISTA's share after excluding ties. The $p$ values use two-sided exact binomial tests over decisive outcomes. Appendix~\ref{app:human} reports the protocol,
category breakdown, and annotator agreement.}
\label{tab:human_main}
\scriptsize
\setlength{\tabcolsep}{3pt}
\renewcommand{\arraystretch}{1.05}
\begin{tabular}{l r >{\columncolor{vistarow}}r r r r}
\toprule
Criterion & Base & \textbf{VISTA} & Tie &
Win$_{\mathrm{dec.}}$ & $p$ \\
\midrule
Compositionality
& 13 & \textbf{65} & 122 & \textbf{83.3\%}
& $1.8{\times}10^{-9}$ \\
Perceived quality
& 45 & 52 & 103 & 53.6\%
& $0.543$ \\
\bottomrule
\end{tabular}
\end{table}

\paragraph{Three scales provide the strongest quality--cost balance.}
The targeted average saturates on both model sizes: a fourth scale raises runtime by $15.0\%$, improves 2B by only $0.004$, and reduces 8B by $0.007$. At the default setting, ImageReward increases by about $20\%$ and CLIPScore by $1.1\%$, whereas the two LAION aesthetic predictors decrease by $1.3$--$2.2\%$ (Appendix~\ref{app:fidelity}). Three scales is therefore our default; reducing the step ceiling to $N_{\max}{=}1$ lowers three-scale runtime to $3.71$\,s. Appendix~\ref{app:overhead} provides the complete runtime and memory breakdown.

\paragraph{Human judgments support compositional improvements across constraint types.}
Annotators evaluated $200$ paired generations presented in randomized order without method identities. As Table~\ref{tab:human_main} shows, VISTA wins $83.3\%$ of decisive pair-level compositionality comparisons ($p<10^{-8}$). The same pattern appears descriptively within every targeted category: VISTA receives $75.5\%$ of decisive annotation-level preferences for attribute binding, $92.8\%$ for 2D spatial relations, and $83.8\%$ for 3D spatial relations. The compositional judgments support the benchmark gains. Separating attention maps may also reduce object and attribute merging, as the qualitative examples illustrate. Annotators show no significant overall preference between Base and VISTA in perceived quality ($p = 0.543$). Appendix~\ref{app:human} reports the complete protocol, counts, category breakdown, and annotator agreement.

%% file: sections/5_limitations.tex
\section{Limitations and Future Work}
\label{sec:limitations}

\textbf{Efficiency and backbone generality.}
VISTA trades additional inference time for compositional control, although Sec.~\ref{sec:scalebudget} shows that this trade-off is adjustable. We evaluate two model sizes, but both belong to the Infinity family. The framework is designed around three general properties of next-scale VAR generation: a continuous input at each scale, persistent state across scales, and discrete sampling; but validation on other VAR backbones remains important. Future work can also reduce overhead through more selective scale and step allocation.

\textbf{Broader compositional coverage remains open.}
We address attribute binding, planar relations, and depth ordering. Numeracy fits the framework but is disabled here, while non-spatial relations require objectives we do not yet instantiate. Complex prompts gain least because they combine several constraints, whereas our objectives treat each constraint separately. Future work can extend the registry with numeracy and non-spatial objectives and develop joint constraints that capture interactions among compositional phenomena. Finally, attribute--noun pairs and relations are obtained from a dependency parse that handles the benchmark prompts and longer examples we examined, but more robust parsing is needed for linguistically diverse prompts.

%% file: sections/6_conclusion.tex
\section{Conclusion}
\label{sec:conclusion}

\textbf{VISTA brings gradient-based test-time alignment to next-scale visual autoregressive generation.}
It separates general-purpose optimization mechanisms from a pluggable registry of compositional objectives, allowing constraints to change without modifying the underlying framework. Instantiated for attribute binding, planar relations, and depth ordering, VISTA improves every targeted T2I-CompBench category on both model sizes and produces large gains in GenEval positioning and attribute binding. Human evaluation confirms stronger compositionality without a significant change in perceived quality. On T2I-CompBench, VISTA also enables a 2B-parameter model to surpass a model four times its size on most targeted categories and on their average.

\textbf{Cross-attention provides a transferable interface for test-time alignment.}
VISTA shows that the central premise of diffusion-based alignment --- using a frozen model's cross-attention as a differentiable, training-free handle on compositional structure --- extends to stateful, multi-scale, discrete-token generation. Whereas existing VAR test-time methods select among sampled trajectories and depend on a satisfying candidate appearing in the sample set, VISTA modifies the trajectory itself toward the requested constraints. By making attention-based objectives usable inside VAR sampling, VISTA opens next-scale autoregressive generators to a broad class of training-free compositional interventions.

%% file: sections/8_1_relatedworks.tex
\section{Related Work}
\label{app:relatedwork}

Sec.~\ref{sec:introduction} positions VISTA between two families of test-time methods: gradient-based alignment for diffusion and inference-time search or guidance for autoregressive generation. We review both more fully here and identify the prior methods underlying our instantiated objectives.

\paragraph{Diffusion methods establish cross-attention as a test-time control interface.}
Attend-and-Excite~\cite{chefer2023attendexcite} optimizes the diffusion latent during sampling to strengthen attention to neglected subject tokens. SynGen~\cite{rassin2023syngen} uses a syntactic parse to align attributes with their nouns while separating them from unrelated concepts. CONFORM~\cite{meral2023conform} casts this alignment as a multi-positive contrastive objective over attention-map embeddings; we use this formulation for attribute binding. Divide\&Bind~\cite{li2023dividebind} adds an attendance objective for multi-instance prompts, while A-STAR~\cite{agarwal2023astar} discourages overlap between concept maps, providing the segregation principle behind our separation term. For planar relations, PSG~\cite{rezaei2025psg} introduces the Probability-of-Superiority objective used by VISTA, while ~\cite{han2025spatial} reposition attention mass through optimal transport.

\paragraph{Diffusion steering does not directly fit next-scale VAR generation.}
These methods share VISTA's premise that a frozen model's cross-attention provides a differentiable, training-free handle on compositional structure. However, they optimize a continuous, fixed-resolution latent that is repeatedly updated across denoising steps. Next-scale VAR models instead generate discrete tokens over increasingly fine grids, with each scale's representation derived and consumed once while persistent state is carried forward. Secs.~\ref{sec:capture}--\ref{sec:budget} provide the optimization mechanisms needed to use the same attention-based objectives in this setting.

\paragraph{Autoregressive test-time methods search or guide.}
TTS-VAR~\cite{chen2025ttsvar} casts next-scale generation as a path-search problem, combining clustering-based diversity preservation at coarse scales with reward-based selection at finer scales. GridAR~\cite{park2026progress} progressively searches and prunes partial raster-scan generations, while ScalingAR~\cite{chen2025go} allocates test-time computation in next-token generation using token-entropy signals. A complementary line modifies next-scale logits without candidate search: SSG~\cite{shin2026ssg} amplifies scale-specific semantic residuals obtained from a frequency-derived coarse prior, whereas VPG~\cite{liao2026vpg} contrasts predictions under genuine and corrupted visual prefixes to reduce prefix drift. These methods target candidate quality, scale consistency, or prefix support, but do not explicitly optimize prompt-derived binding and spatial constraints. VISTA instead differentiates compositional objectives through cross-attention to modify a single coarse-scale representation. Table~\ref{tab:main} compares VISTA directly with TTS-VAR and BoN, and Appendix~\ref{app:arguidance} gives a protocol-aware comparison with SSG and VPG.

\paragraph{Depth ordering can be approached through different spatial cues.}
For 3D relations, PSG~\cite{rezaei2025psg} searches over completed images using segmentation and monocular-depth estimators. LaRender~\cite{zhan2025larender} instead applies volume-rendering principles in diffusion latent space using object regions and an occlusion order. Our proposed depth objective provides a complementary attention-only formulation based on the prompt relation and the backbone's cross-attention, without requiring a depth estimator or supplied layout. It is one objective instantiated within VISTA rather than a requirement of the framework.

\paragraph{VISTA builds on next-scale visual autoregressive generation.}
VAR~\cite{tian2024var} introduces coarse-to-fine next-scale prediction, and Infinity~\cite{han2024infinity} extends it to text-conditioned generation with a bitwise tokenizer and self-correction. Prior benchmarking work~\cite{shahabadi2025infinitybeyond} documents Infinity's compositional failures relative to diffusion and diffusion-transformer systems and provides the prior-system results in Table~\ref{tab:main}. ScaleKV~\cite{li2025scalekv} reduces Infinity's cache cost by allocating different KV budgets across layers and scales. We apply it to all Infinity configurations, with Appendix~\ref{app:seeds} showing that its effect on the reported compositional scores is negligible.

\paragraph{The benchmarks expose complementary aspects of compositionality.}
T2I-CompBench~\cite{huang2023t2icompbench} evaluates attribute binding, spatial and non-spatial relations, counting, and complex compositions through category-specific prompts and metrics. Its separation of constraint types lets individual categories isolate the corresponding VISTA objectives. GenEval~\cite{ghosh2023geneval} provides a complementary object-centric evaluation of presence, counting, color attribution, attribute binding, and relative position. Together, they expose recurring failures across model families, including attribute leakage, missing objects, incorrect arrangements, and miscounting.

%% file: sections/8_2_appendix.tex
\section{Objective Definitions and Provenance}
\label{app:borrowed}

This appendix specifies the binding and planar objectives used by VISTA and distinguishes components adopted from prior work from those introduced here. Both objectives operate on the token-group attention map $\mathbf{m}_T$ defined in Sec.~\ref{sec:objectives}.

\paragraph{Attribute binding.}
Following attention-based binding methods~\cite{rassin2023syngen,meral2023conform}, let
$p=(a,n)\in\mathcal{P}$ denote an attribute--noun pair, where $a$ and $n$ are the token groups corresponding to the modifier and its noun. For example, ``a red banana'' gives
$p=(\textsf{red},\textsf{banana})$.
For each $T\in\{a,n\}$, we reshape $\mathbf{m}_T$ to the spatial grid, smooth it with a fixed $3{\times}3$ Gaussian kernel $G$ with $\sigma=0.5$, and $\ell_2$-normalize the flattened result:
\begin{equation}
\mathbf{u}_T =
\frac{
\operatorname{vec}\!\left(
G * \operatorname{reshape}(\mathbf{m}_T)
\right)}
{
\left\|
\operatorname{vec}\!\left(
G * \operatorname{reshape}(\mathbf{m}_T)
\right)
\right\|_2
}.
\end{equation}

The resulting $M=2|\mathcal{P}|$ embeddings form a contrastive objective adapted from CONFORM~\cite{meral2023conform}. For an embedding $i$, $P(i)$ contains the embedding associated with the other member of the same attribute--noun pair, while embeddings from other pairs act as negatives. With
$\mathrm{sim}_{ij}=\mathbf{u}_i^\top\mathbf{u}_j/\eta$
and temperature $\eta=0.07$,
\begin{equation}
\mathcal{L}_{\mathrm{bind}}(\mathcal{P})
=
-\frac{1}{M}
\sum_{i=1}^{M}
\frac{1}{|P(i)|}
\sum_{j\in P(i)}
\log
\frac{
\exp(\mathrm{sim}_{ij})
}{
\sum_{j'\neq i}\exp(\mathrm{sim}_{ij'})
}.
\label{eq:bind}
\end{equation}
The loss is evaluated jointly over all extracted pairs, encouraging each attribute map to align with its corresponding noun while remaining distinguishable from competing pairs.

\paragraph{Planar relations.}
For a relation
$r=(X,\rho,Y)\in\mathcal{R}_{\mathrm{2D}}$
along image-plane axis $a$, we use the probability-of-superiority formulation from PSG~\cite{rezaei2025psg}. Each attention map is first converted into a spatial distribution,
\begin{equation}
p_Z = \operatorname{softmax}(\gamma\mathbf{m}_Z),
\qquad Z\in\{X,Y\},
\end{equation}
where $\gamma$ sharpens the diffuse attention maps encountered at coarse VAR scales. The probability that $X$ lies above $Y$ in the ordering induced by axis $a$ is
\begin{equation}
\operatorname{PoS}_a(X,Y)
=
\sum_{i,j}
p_X(i)\,p_Y(j)\,
\mathbb{1}\!\left[
\operatorname{coord}_a(i)
\geq
\operatorname{coord}_a(j)
\right].
\label{eq:pos}
\end{equation}
For relations aligned with the increasing coordinate direction, we set
$q_r=\operatorname{PoS}_a(X,Y)$; for the opposite direction, such as \emph{left of} or \emph{above}, we use
$q_r=1-\operatorname{PoS}_a(X,Y)$.
The planar objective is
\begin{equation}
\mathcal{L}_{\mathrm{2D}}
=
-\frac{1}{|\mathcal{R}_{\mathrm{2D}}|}
\sum_{r\in\mathcal{R}_{\mathrm{2D}}}
q_r^2,
\end{equation}
where the probability-of-superiority form and its squared objective follow PSG.

\paragraph{Provenance.}
Table~\ref{tab:provenance} summarizes the origin of each objective component. We distinguish adopting a motivating principle from adopting a functional form.

\begin{table}[H]
\centering
\caption{Provenance of the objective terms used in VISTA.}
\label{tab:provenance}
\small
\setlength{\tabcolsep}{4pt}
\begin{tabular}{llp{6.1cm}p{3.4cm}}
\toprule
Term & Eq. & Prior work & VISTA-specific component \\
\midrule
$\mathcal{L}_{\mathrm{bind}}$
& \ref{eq:bind}
& Contrastive binding formulation adapted from CONFORM~\cite{meral2023conform}
& Application to grouped attention maps at coarse VAR scales \\

$\mathcal{L}_{\mathrm{2D}}$
& \ref{eq:pos}
& Probability-of-superiority formulation and squared objective from PSG~\cite{rezaei2025psg}
& Temperature sharpening and directional implementation on discrete VAR grids \\

$\mathcal{L}_{\mathrm{sep}}$
& \ref{eq:sep}
& Segregation principle motivated by A-STAR~\cite{agarwal2023astar}
& Extent-based functional form \\

$\mathcal{L}_{\mathrm{bc}}$
& \ref{eq:bc}
& ---
& Boundary-containment objective \\
\midrule
$\mathcal{L}_{\mathrm{entity}}$
& ---
& ---
& Not instantiated \\

$\mathcal{L}_{\mathrm{ns}}$
& ---
& ---
& Not instantiated \\
\bottomrule
\end{tabular}
\end{table}

We claim no new objective formulation for attribute binding or planar relations. The objective-level contribution is the depth term
$\mathcal{L}_{\mathrm{depth}}
=\mathcal{L}_{\mathrm{sep}}+\mathcal{L}_{\mathrm{bc}}$:
$\mathcal{L}_{\mathrm{sep}}$ adopts the segregation motivation of A-STAR but uses a different extent-based function, while
$\mathcal{L}_{\mathrm{bc}}$ is introduced in this work. Unlike approaches that evaluate completed candidates using segmentation and monocular-depth models, this objective extracts a differentiable occlusion cue directly from the backbone's cross-attention maps and requires no auxiliary model.

\section{Depth Objective: Construction Details}
\label{app:bac}

This section specifies the depth objective summarized in Sec.~\ref{sec:objectives}. For a relation ``$B$ behind $A$,'' $A$ denotes the foreground object and $B$ the occluded object. We convert their attention maps into spatial distributions,
\begin{equation}
p_X=\operatorname{softmax}(\gamma\mathbf{m}_X),
\qquad X\in\{A,B\},
\end{equation}
defined over normalized grid coordinates
$\mathbf{c}_i\in[0,1]^2$.
Because cross-attention provides no viewing-axis coordinate, the objective uses overlap and boundary-containment cues associated with occlusion rather than estimating metric depth.

\paragraph{Separation.}
For each axis $a\in\{x,y\}$, we compute the attention-weighted centroid and spread
\begin{equation}
\mu_a^X
=
\sum_i p_X(i)c_{i,a},
\qquad
\varrho_a^X
=
\sqrt{
\sum_i p_X(i)
\bigl(c_{i,a}-\mu_a^X\bigr)^2
}.
\end{equation}
The separation term penalizes the intersection of the resulting axis-aligned extents:
\begin{equation}
\mathcal{L}_{\mathrm{sep}}
=
\prod_{a\in\{x,y\}}
\left[
\min_{X\in\{A,B\}}
\bigl(\mu_a^X+\varrho_a^X\bigr)
-
\max_{X\in\{A,B\}}
\bigl(\mu_a^X-\varrho_a^X\bigr)
\right]_+ ,
\label{eq:sep}
\end{equation}
where $[u]_+=\max(u,0)$. This discourages the two entities from collapsing into the same attention support.

\paragraph{Facing boundary.}
Separation alone does not assign foreground--background order. We approximate the perimeter of $B$ using the spatial gradient magnitude of its reshaped attention distribution:
\begin{equation}
e_B(i)
=
\sqrt{
\bigl(D_xp_B(i)\bigr)^2+
\bigl(D_yp_B(i)\bigr)^2
}.
\end{equation}
The derivatives use central finite differences in the grid interior and one-sided differences at its boundaries. We then emphasize the portion of this perimeter facing $A$:
\begin{equation}
s_i
=
e_B(i)
\exp\!\left(
-\frac{
\left\|
\mathbf{c}_i-\operatorname{sg}(\boldsymbol{\mu}^{A})
\right\|_2
}{\kappa}
\right),
\qquad
\boldsymbol{\mu}^{A}=(\mu_x^A,\mu_y^A),
\end{equation}
where $\operatorname{sg}$ denotes stop-gradient. This prevents the facing-selection term from being reduced simply by moving the centroid of $A$.

\paragraph{Virtual boundary points.}
Let $\mathcal{I}_K$ contain the $K$ locations with the largest $s_i$, or all locations when the grid contains fewer than $K$, and define
\begin{equation}
w_k=
\frac{s_k}{\sum_{\ell\in\mathcal{I}_K}s_\ell}.
\end{equation}
For
$\boldsymbol{\alpha}^{(v)}
\sim\operatorname{Dir}
(\mathbf{1}_{|\mathcal{I}_K|})$,
we construct
\begin{equation}
\widetilde{\mathbf{x}}_v
=
\sum_{k\in\mathcal{I}_K}
\frac{
\alpha_k^{(v)}w_k
}{
\sum_{\ell\in\mathcal{I}_K}
\alpha_\ell^{(v)}w_\ell
}
\mathbf{c}_k,
\qquad v=1,\ldots,V.
\end{equation}
These virtual points lie in the convex hull of the selected boundary locations and concentrate on the portion of $B$ facing $A$.

\paragraph{Boundary containment.}
We evaluate the attention density of $A$ at each virtual point using a Gaussian kernel:
\begin{equation}
d_A(\widetilde{\mathbf{x}}_v)
=
\sum_i p_A(i)
\exp\!\left(
-\frac{
\left\|
\mathbf{c}_i-\widetilde{\mathbf{x}}_v
\right\|_2^2
}{2h^2}
\right).
\end{equation}
The containment loss requires these points to fall within sufficiently strong support from the foreground object:
\begin{equation}
\mathcal{L}_{\mathrm{bc}}
=
\frac{1}{V}
\sum_{v=1}^{V}
\left[
\delta-d_A(\widetilde{\mathbf{x}}_v)
\right]_+ .
\label{eq:bc}
\end{equation}
We set $\delta$ relative to the peak attention of $A$, adapting the containment threshold to the concentration of its map.

\paragraph{Combined objective.}
The complete objective is
\begin{equation}
\mathcal{L}_{\mathrm{depth}}
=
\lambda_{\mathrm{sep}}\mathcal{L}_{\mathrm{sep}}
+
\lambda_{\mathrm{bc}}\mathcal{L}_{\mathrm{bc}}.
\end{equation}
The separation term preserves distinct entity supports, while boundary containment assigns their foreground--background order through an occlusion cue. The objective does not recover metric depth or a general 3D representation.

All experiments use
$\gamma=8$,
$K=64$,
$V=256$,
$\kappa=0.3$,
$h=0.15$,
$\delta=0.3\max_i p_A(i)$, and
$\lambda_{\mathrm{sep}}=\lambda_{\mathrm{bc}}=1$.
Alternative depth objectives under the same VISTA machinery are evaluated in Appendix~\ref{app:objvariants}.

\section{Adaptive Step-Budget Scheduler}
\label{app:budget}

For each objective active at the initial step, we define
\begin{equation}
\varsigma_k =
\operatorname{clip}\!\left(
\frac{\log\!\bigl(\overline{g}^{(k,0)}/\tau_k\bigr)}
{\log(N_{\max}+1)},
\,0,\,1
\right),
\qquad
N_{\mathrm{eff}} =
\max\!\left(
1,
\left\lceil
N_{\max}\max_{k\in\mathcal{K}_0}\varsigma_k
\right\rceil
\right).
\label{eq:severity}
\end{equation}

The clipped log ratio maps the initial constraint violation to a bounded step budget while reducing sensitivity to large differences in gradient magnitude. The maximum allocates the budget according to the most violated active objective, and the outer maximum ensures that every selected scale receives at least one update. $N_{\mathrm{eff}}$ is only an upper bound: optimization terminates earlier when all objectives fall below their activity thresholds. We use this scheduler as a fixed compute-control heuristic throughout the experiments.

\section{Alternative Objective Instantiations}
\label{app:objvariants}

VISTA separates its optimization machinery from the objectives applied to the captured cross-attention maps. To test this interface, we hold the capture, normalization, combination, and budget mechanisms fixed and substitute alternative binding, planar, and depth objectives.

Each instantiation is evaluated with one seed on a fixed subset of $100$ prompts from its corresponding T2I-CompBench category. Color, Texture, and Shape isolate the binding objective; 2D Spatial isolates the planar objective; and 3D Spatial isolates the depth objective. Complex prompts are omitted because they activate multiple terms simultaneously. The unsteered baselines are recomputed on the same subsets, so these diagnostic results should not be compared row by row with the full-benchmark results in Table~\ref{tab:main}. The experiment tests compatibility with the shared interface rather than providing a statistically powered ranking of objectives.

\begin{table}[H]
\centering
\caption{Binding objectives on the three categories where only $\mathcal{L}_\mathrm{bind}$ is active. All rows share the same optimization machinery and the same planar and depth terms, which are inactive here. $\Delta$ is on the mean.}
\label{tab:objbind}
\small
\setlength{\tabcolsep}{5pt}
\begin{tabular}{lcccc|c}
\toprule
$\mathcal{L}_\mathrm{bind}$ & Color & Texture & Shape & Mean & $\Delta$ \\
\midrule
\rowcolor{vistarow} CONFORM (default) & 0.830 & 0.764 & 0.570 & 0.721 & $+17.1\%$ \\
SynGen divergence & 0.823 & 0.745 & 0.565 & 0.711 & $+15.5\%$ \\
KL & 0.830 & 0.740 & 0.556 & 0.709 & $+15.1\%$ \\
Focal & 0.769 & 0.735 & 0.579 & 0.694 & $+12.8\%$ \\
Divide-and-Bind & 0.778 & 0.633 & 0.459 & 0.623 & $+1.2\%$ \\
JSD & 0.730 & 0.642 & 0.475 & 0.616 & $+0.0\%$ \\
\midrule
No steering & 0.730 & 0.642 & 0.475 & 0.616 & --- \\
\bottomrule
\end{tabular}
\end{table}

\begin{table}[H]
\centering
\begin{minipage}[t]{0.48\textwidth}
\centering
\caption{Planar objectives on 2D Spatial, where only $\mathcal{L}_\mathrm{2D}$ is active.}
\label{tab:obj2d}
\small
\begin{tabular}{lcc}
\toprule
$\mathcal{L}_\mathrm{2D}$ & 2D Spatial & $\Delta$ \\
\midrule
Relative centroid & 0.473 & $+175.9\%$ \\
\rowcolor{vistarow} PoS (default) & 0.454 & $+164.8\%$ \\
Gaussian-at-target & 0.417 & $+143.3\%$ \\
Half-field mass & 0.363 & $+111.8\%$ \\
Separation & 0.225 & $+31.3\%$ \\
Attention IoU & 0.198 & $+15.5\%$ \\
\midrule
No steering & 0.171 & --- \\
\bottomrule
\end{tabular}
\end{minipage}\hfill
\begin{minipage}[t]{0.48\textwidth}
\centering
\caption{Depth objectives on 3D Spatial, where only $\mathcal{L}_\mathrm{depth}$ is active.}
\label{tab:obj3d}
\small
\begin{tabular}{lcc}
\toprule
$\mathcal{L}_\mathrm{depth}$ & 3D Spatial & $\Delta$ \\
\midrule
\rowcolor{vistarow} BC (ours) & 0.495 & $+12.8\%$ \\
Peak dominance & 0.487 & $+11.0\%$ \\
Overlap dominance & 0.478 & $+8.9\%$ \\
Containment--adjacency & 0.458 & $+4.3\%$ \\
Depth proxy & 0.419 & $-4.5\%$ \\
\midrule
No steering & 0.439 & --- \\
\bottomrule
\end{tabular}
\end{minipage}
\end{table}

Across the $17$ tested instantiations, $15$ improve over their corresponding unsteered baseline, one matches it, and one underperforms. Thus, objectives with substantially different formulations can use the same capture and optimization machinery without framework-specific modifications. This supports VISTA's objective-agnostic interface; it does not imply that every possible attention objective will be equally effective.

A descriptive pattern is that objectives operating on spatial summaries transfer more reliably at coarse scales than objectives based primarily on elementwise map comparisons. Boundary containment performs best among the tested depth objectives. For planar relations, relative centroid slightly exceeds the literature-derived PoS objective on this diagnostic subset. We retain PoS as the default to keep the main configuration tied to the established objective rather than selecting a replacement retrospectively on this small study. Given the single seed and subset size, we draw no conclusions from small differences between the stronger variants.

\section{Per-Seed Results and Variance}
\label{app:seeds}

Table~\ref{tab:seeds} reports per-seed T2I-CompBench results for VISTA's default configuration on both backbones. VISTA and the corresponding Infinity baselines in Table~\ref{tab:main} are averaged over four seeds. BoN and TTS-VAR use one seed, while results for other systems are taken from their respective papers.

\begin{table}[H]
\centering
\caption{Per-seed T2I-CompBench scores for VISTA's default configuration of three steered scales and $N_{\max}{=}5$. Avg.\ is the mean over the six targeted categories.}
\label{tab:seeds}
\footnotesize
\setlength{\tabcolsep}{2.5pt}
\begin{tabular}{llccccccc}
\toprule
Model & Seed & Col. & Tex. & Shp. & 2D & 3D & Cpx. & Avg. \\
\midrule
\multirow{5}{*}{2B}
& 0 & 0.826 & 0.750 & 0.566 & 0.458 & 0.455 & 0.394 & 0.575 \\
& 1 & 0.823 & 0.755 & 0.579 & 0.432 & 0.446 & 0.400 & 0.573 \\
& 2 & 0.853 & 0.744 & 0.576 & 0.439 & 0.453 & 0.399 & 0.577 \\
& 3 & 0.828 & 0.752 & 0.576 & 0.438 & 0.453 & 0.399 & 0.574 \\
\cmidrule(l){2-9}
& Std. & 0.014 & 0.005 & 0.006 & 0.011 & 0.004 & 0.003 & 0.002 \\
\midrule
\multirow{5}{*}{8B}
& 0 & 0.851 & 0.802 & 0.652 & 0.429 & 0.426 & 0.406 & 0.594 \\
& 1 & 0.863 & 0.790 & 0.666 & 0.417 & 0.424 & 0.402 & 0.594 \\
& 2 & 0.858 & 0.801 & 0.657 & 0.404 & 0.421 & 0.402 & 0.591 \\
& 3 & 0.860 & 0.800 & 0.656 & 0.411 & 0.425 & 0.402 & 0.592 \\
\cmidrule(l){2-9}
& Std. & 0.005 & 0.006 & 0.006 & 0.011 & 0.002 & 0.002 & 0.002 \\
\bottomrule
\end{tabular}
\end{table}

\paragraph{Variance across seeds.}
The targeted-average standard deviation is $0.002$ on both backbones, and the largest category-level standard deviation is $0.014$. The improvements in the principal targeted categories therefore remain stable across seeds, including the large gain on 2D Spatial. Complex has the smallest improvement and should be interpreted more cautiously; our conclusions do not rely on that category alone.

\paragraph{Effect of cache compression.}
All Infinity results in Table~\ref{tab:main} use ScaleKV, since VISTA is applied on top of the compressed backbone. As a reference check, the uncompressed Infinity-2B baseline obtains a targeted average of $0.481$, compared with $0.480$ with ScaleKV, and no category changes by more than $0.008$. The effect is comparable to the observed seed variation and negligible at the level of the reported aggregate results. We therefore use the compressed backbone consistently throughout.

\section{Additional Qualitative Results}
\label{app:qualitative}

\begin{figure*}[t]
\centering
\setlength{\tabcolsep}{2pt}
\renewcommand{\arraystretch}{0.4}
\begin{tabular}{
@{}c@{\hspace{2pt}}
*{4}{>{\centering\arraybackslash}p{0.215\textwidth}}
@{}
}
&
{\scriptsize ``a cat \textcolor{depthorange}{in front of} a dog''}
&
{\scriptsize ``a bird \textcolor{depthorange}{in front of} a dog''}
&
{\scriptsize ``a bowl \textcolor{depthorange}{behind} a banana''}
&
{\scriptsize ``a table \textcolor{depthorange}{behind} a dog''}
\\[1pt]
\rotatebox{90}{\scriptsize\textbf{Base}}
&
\includegraphics[width=\linewidth]{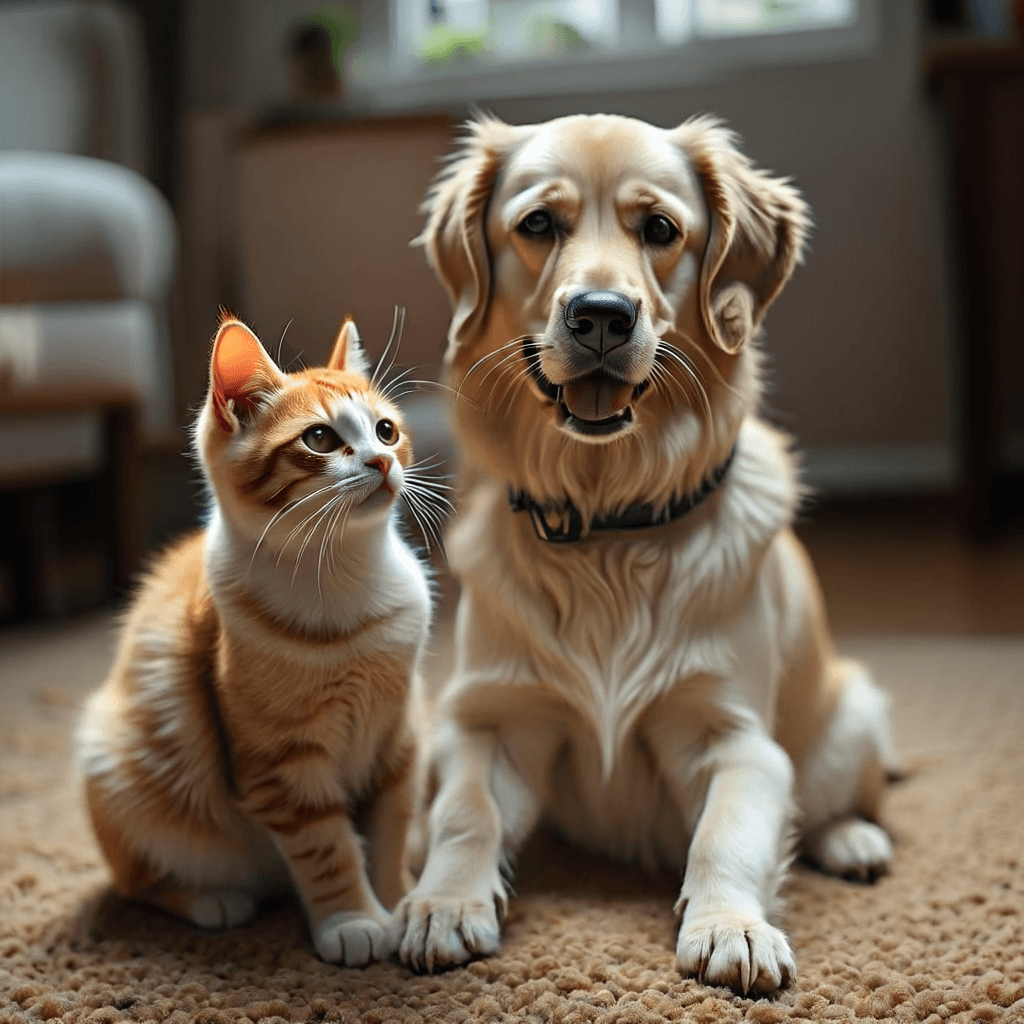}
&
\includegraphics[width=\linewidth]{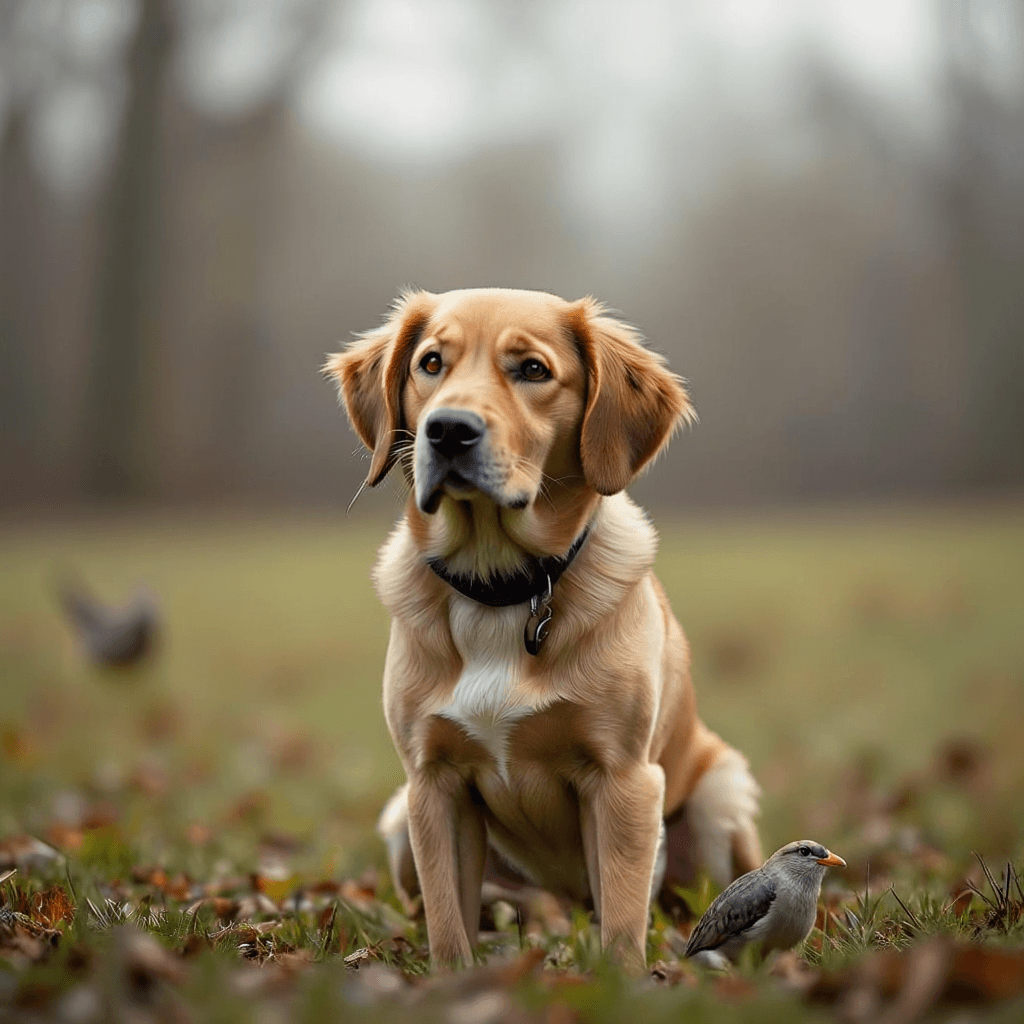}
&
\includegraphics[width=\linewidth]{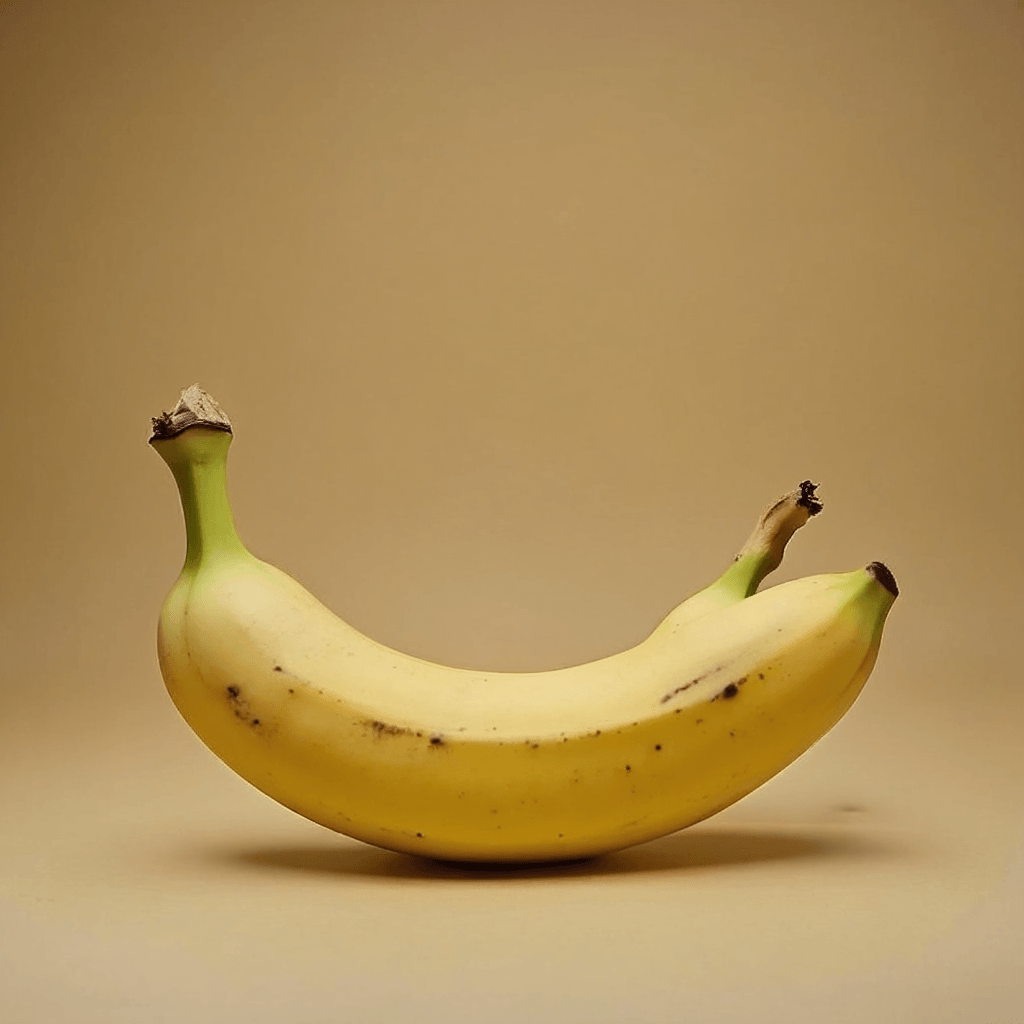}
&
\includegraphics[width=\linewidth]{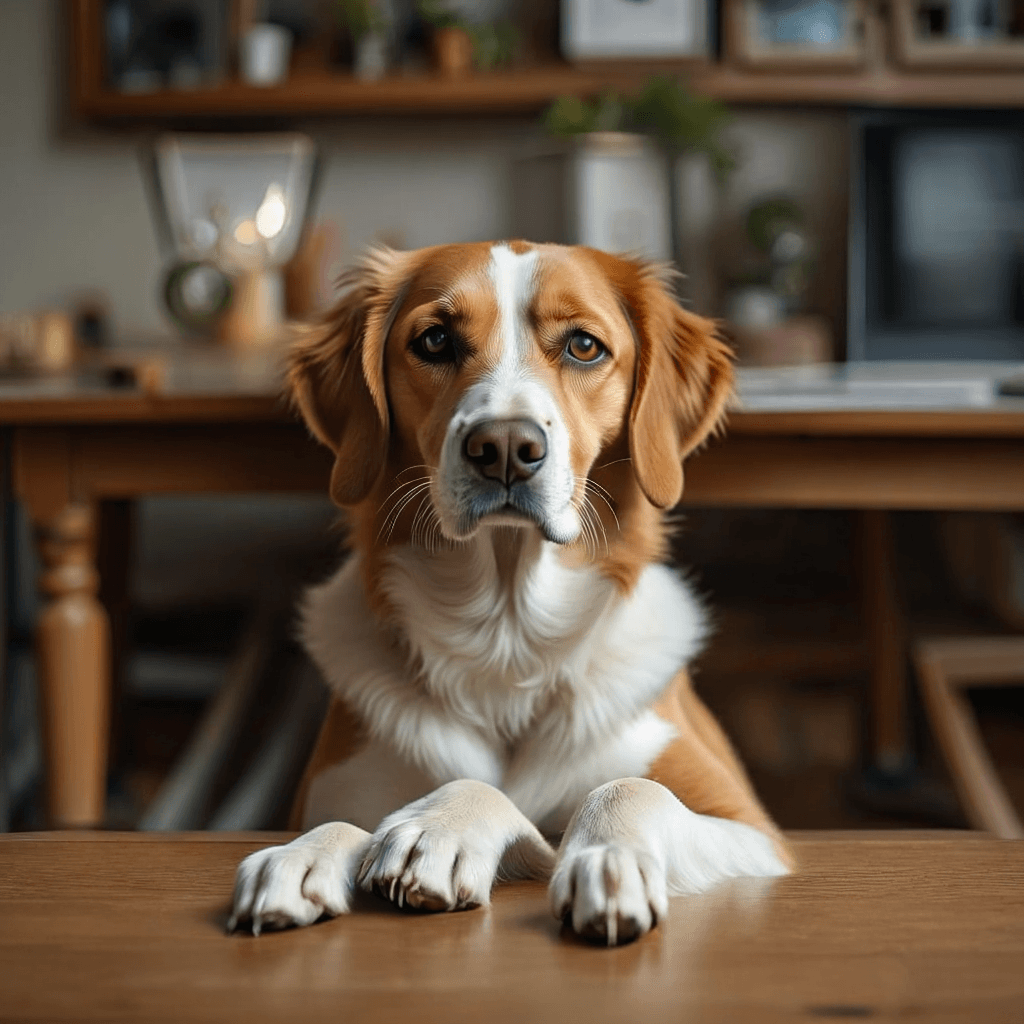}
\\[1pt]
\rotatebox{90}{\scriptsize\textbf{VISTA (Ours)}}
&
\includegraphics[width=\linewidth]{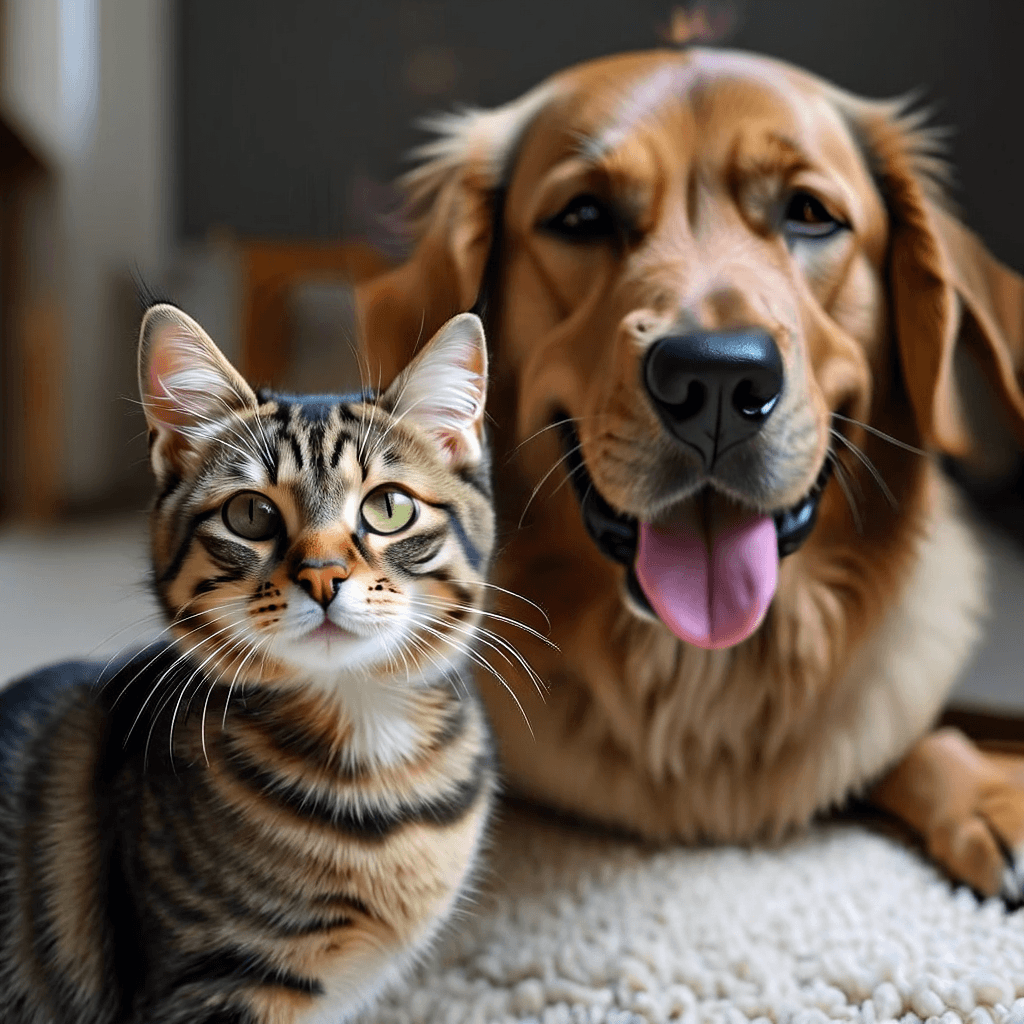}
&
\includegraphics[width=\linewidth]{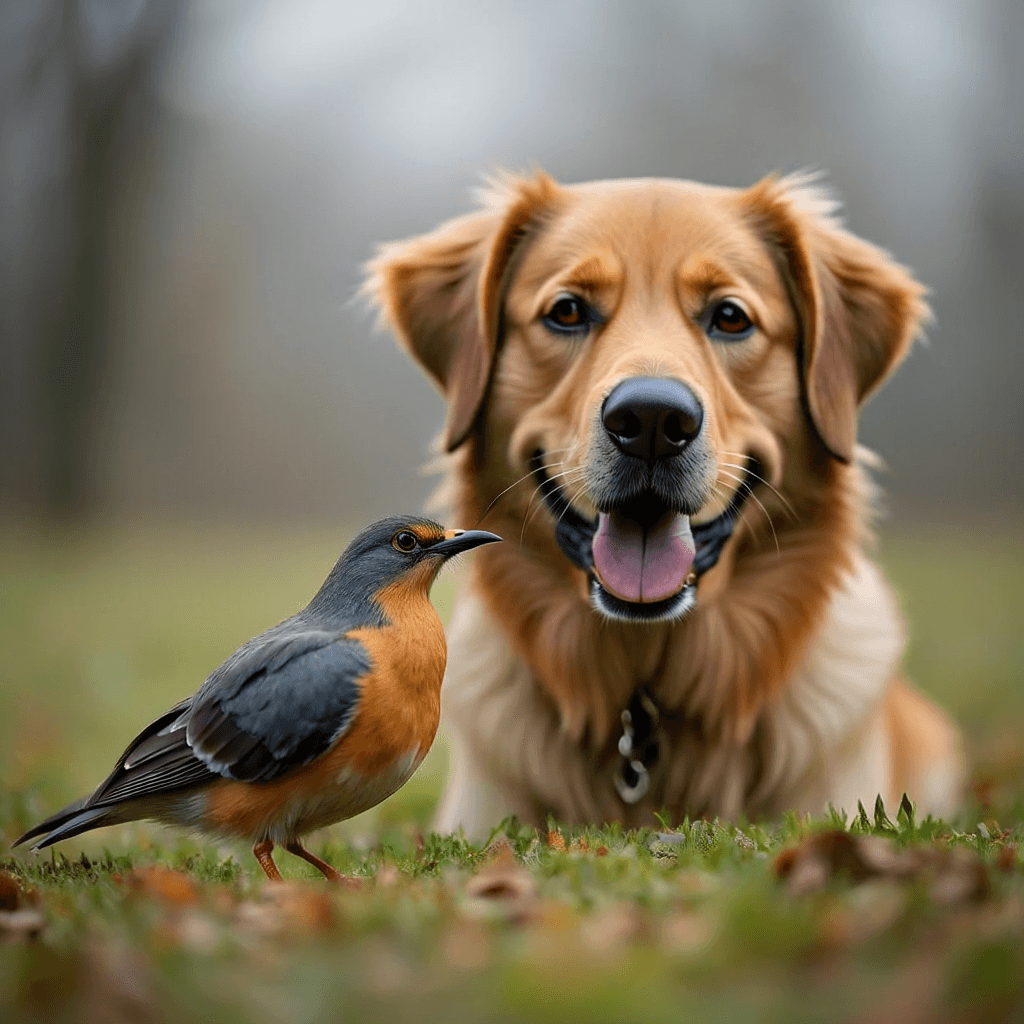}
&
\includegraphics[width=\linewidth]{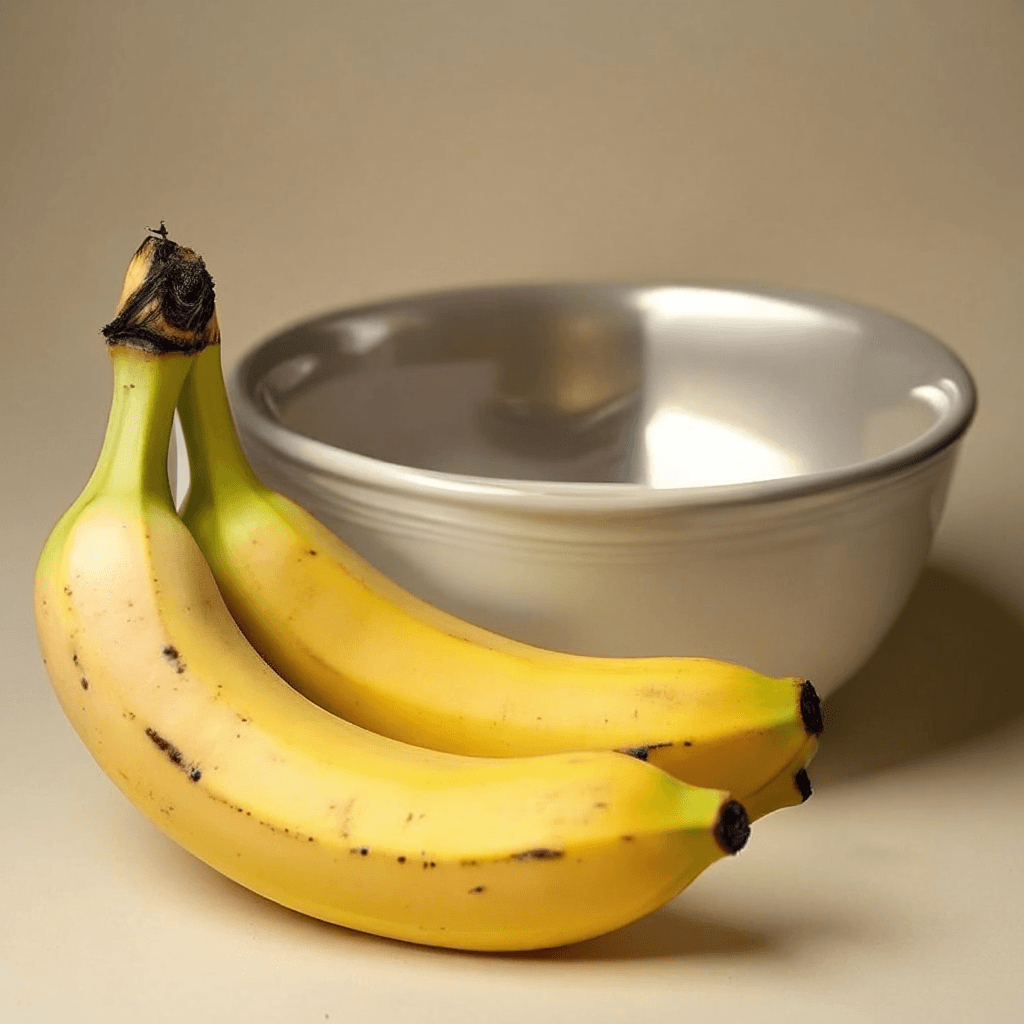}
&
\includegraphics[width=\linewidth]{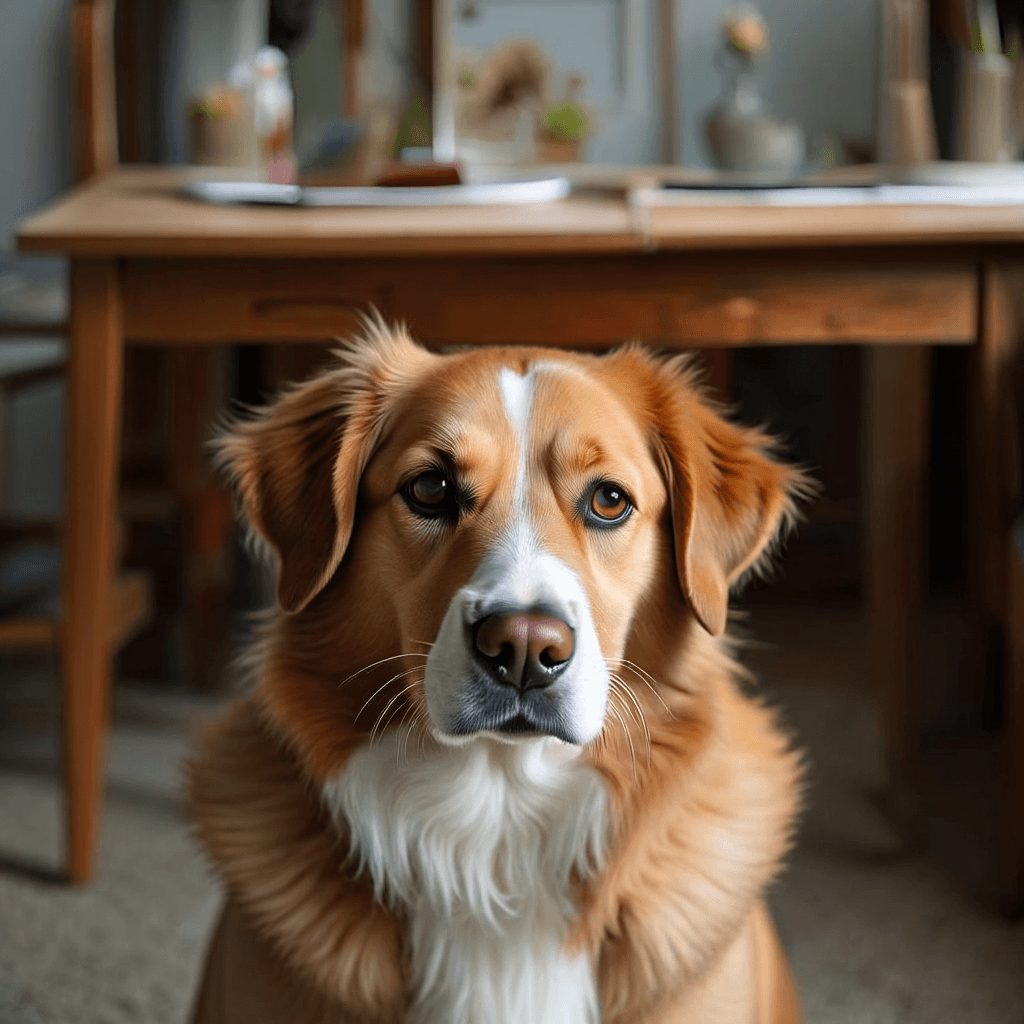}
\\
\end{tabular}
\caption{Selected qualitative comparisons for depth ordering. Each column uses the same prompt and random seed for Base and VISTA. The prompts contain only a front--back relation, isolating $\mathcal{L}_{\mathrm{depth}}$ from the binding and planar objectives. Across these examples, VISTA produces clearer overlap and occlusion cues consistent with the requested ordering. Quantitative results over the complete 3D Spatial category appear in Table~\ref{tab:main}.}
\label{fig:qualitative_depth}
\end{figure*}

Unlike planar relations, front--back order cannot be expressed by ordering two objects along an image-plane axis. The depth objective instead encourages separation and boundary-containment cues that make the requested ordering visible through occlusion. Figure~\ref{fig:qualitative_depth} illustrates this intended behavior; it does not assume that cross-attention encodes metric depth.

\section{Human Evaluation}
\label{app:human}

Four annotators independently evaluated $200$ paired Base--VISTA
generations: $100$ attribute-binding, $50$ 2D-spatial, and $50$
3D-spatial prompts sampled from T2I-CompBench. Image order was randomized independently for each comparison, and annotators were unaware of the method identities. Each annotator made
separate three-way judgments (VISTA, Base, or tie) for compositional
faithfulness and visual quality. For each pair and criterion, we select the label with the highest vote count among VISTA, Base, and tie; equal highest counts produce a pair-level tie. Thus, a $2$--$1$--$1$ split selects the two-vote label, while a $2$--$2$ split produces a tie. We report annotation-level category results descriptively and use two-sided exact binomial tests over decisive pair-level outcomes.

\begin{table}[H]
\centering
\caption{Human evaluation. Win denotes VISTA's share among decisive
judgments. Category rows aggregate four annotations per pair; pair-level rows contain one plurality-aggregated outcome per image pair.}
\label{tab:human}
\small
\setlength{\tabcolsep}{5pt}
\begin{tabular}{lrrrrr}
\toprule
Criterion & $n$ & VISTA & Base & Tie & Win \\
\midrule
\multicolumn{6}{l}{\emph{Compositional faithfulness}} \\
Attribute binding & 400 & 145 & 47 & 208 & 75.5\% \\
3D spatial        & 200 & 57  & 11 & 132 & 83.8\% \\
2D spatial        & 200 & 90  & 7  & 103 & 92.8\% \\
All annotations   & 800 & 292 & 65 & 443 & 81.8\% \\
Pair-level vote     & 200 & 65  & 13 & 122 & 83.3\% \\
\midrule
\multicolumn{6}{l}{\emph{Visual quality}} \\
All annotations   & 800 & 238 & 218 & 344 & 52.2\% \\
Pair-level vote     & 200 & 52  & 45  & 103 & 53.6\% \\
\bottomrule
\end{tabular}
\end{table}

Agreement was moderate for compositional faithfulness
(Fleiss' $\kappa=0.589$) and fair for visual quality
($\kappa=0.333$). For decisive pair-level outcomes, VISTA's compositional advantage is significant ($p=1.8\times10^{-9}$), whereas visual quality shows no significant preference ($p=0.543$).

\section{Prompt Fidelity and Visual Quality}
\label{app:fidelity}

We evaluate Infinity-2B and VISTA on $600$ T2I-CompBench prompts, with $100$ prompts from each targeted category. The random seed is held fixed, pairing every VISTA generation with a base generation from the same prompt and seed. We report three metric families that are not used by VISTA: ImageReward~\cite{xu2023imagereward}, CLIPScore~\cite{hessel2021clipscore}, and two released versions of the LAION aesthetic predictor~\cite{schuhmann2022laion}. Statistical significance is assessed using a two-sided Wilcoxon signed-rank test over the $600$ paired generations. This single-seed analysis complements the multi-seed compositional results in Table~\ref{tab:main}.

\begin{table}[H]
\centering
\caption{Prompt-conditioned and aesthetic scores as the number of steered scales increases. All VISTA generations are paired with base generations using the same prompts and seeds. Differences and $p$ values compare the default three-scale configuration with the base model.}
\label{tab:fidelity}
\small
\setlength{\tabcolsep}{5pt}
\begin{tabular}{lccccccc}
\toprule
Metric & Base & 1 scale & 2 scales & 3 scales & 4 scales
& $\Delta$ at default & $p$ \\
\midrule
ImageReward
& 1.025 & \textbf{1.234} & \textbf{1.234} & 1.230 & 1.204
& $+20.1\%$ & $3.6{\times}10^{-7}$ \\
CLIPScore
& 0.321 & 0.325 & \textbf{0.326} & 0.325 & 0.324
& $+1.1\%$ & $6.1{\times}10^{-4}$ \\
Aesthetic (v1)
& \textbf{6.056} & 6.022 & 6.023 & 5.976 & 5.957
& $-1.3\%$ & $1.6{\times}10^{-3}$ \\
Aesthetic (v2)
& \textbf{5.734} & 5.683 & 5.615 & 5.609 & 5.608
& $-2.2\%$ & $2.0{\times}10^{-10}$ \\
\bottomrule
\end{tabular}
\end{table}

\paragraph{Steering improves prompt-conditioned scores while preserving perceived quality.}
At the default three-scale configuration, ImageReward increases by $20.1\%$ and CLIPScore by $1.1\%$, while the two aesthetic predictors decrease modestly by $1.3\%$ and $2.2\%$. These metric families measure different properties: ImageReward and CLIPScore condition on the prompt, whereas the aesthetic predictors score the image independently of it. The results therefore indicate substantially stronger prompt agreement with a small trade-off in predicted aesthetics. Importantly, the human evaluation in Table~\ref{tab:human_main} finds no significant preference in perceived quality ($p=0.54$), providing no evidence of a human-perceptible quality penalty at the default setting.

\paragraph{The improvement is not driven by a single category.}
At the default configuration, ImageReward increases in all six evaluated categories, with per-category gains ranging from $+0.034$ on Complex to $+0.351$ on Texture. This consistency indicates that the aggregate improvement is distributed across the benchmark rather than arising from one category.

\section{Inference Overhead: Full Breakdown}
\label{app:overhead}

Table~\ref{tab:overhead_full} reports per-image generation time on Infinity-2B over every combination of step budget, steered-scale count, and active objective subset. All 2B timing configurations use the same $20$ prompts and an NVIDIA GeForce RTX 3090 (24\,GB). The Infinity-8B experiments use an NVIDIA A100 (40\,GB). The unmodified baseline averages $3.34$\,s per image. Peak memory is $10.96$\,GB for the baseline and all configurations with up to three steered scales, rising to $15.1$--$15.3$\,GB at four scales.

\begin{table}[H]
\centering
\caption{Per-image generation time in seconds on Infinity-2B. B, S, and D denote the binding, spatial, and depth objectives. The unmodified baseline is $3.34$\,s.}
\label{tab:overhead_full}
\small
\setlength{\tabcolsep}{3.5pt}
\begin{tabular}{lccccccc}
\toprule
Scales & B & S & D & B+S & B+D & S+D & B+S+D \\
\midrule
\multicolumn{8}{l}{\emph{$N_{\max}=1$}} \\
1 & 3.34 & 3.31 & 3.34 & 3.39 & 3.38 & 3.39 & 3.42 \\
2 & 3.44 & 3.45 & 3.45 & 3.54 & 3.50 & 3.52 & 3.57 \\
3 & 3.53 & 3.54 & 3.56 & 3.71 & 3.63 & 3.64 & 3.71 \\
4 & 3.63 & 3.63 & 3.64 & 3.83 & 3.75 & 3.79 & 3.88 \\
\midrule
\multicolumn{8}{l}{\emph{$N_{\max}=5$}} \\
1 & 3.71 & 3.71 & 3.72 & 3.94 & 3.85 & 3.89 & 3.99 \\
2 & 4.17 & 4.14 & 4.16 & 4.63 & 4.46 & 4.51 & 4.74 \\
3 & 4.62 & 4.48 & 4.48 & 5.28 & 5.06 & 4.92 & 5.46 \\
4 & 5.11 & 4.61 & 4.62 & 6.03 & 5.74 & 5.08 & 6.28 \\
\bottomrule
\end{tabular}
\end{table}

\paragraph{Objectives compose sub-additively.}
At the default setting of three scales and $N_{\max}{=}5$, binding, spatial, and depth steering add $1.28$\,s, $1.14$\,s, and $1.14$\,s when run independently. Summing these standalone costs would add $3.56$\,s, whereas jointly optimizing all three adds $2.12$\,s, approximately $60\%$ as much. This follows from Sec.~\ref{sec:combine}: the active objectives share the same optimization passes and contribute to one joint update rather than running sequentially.

\paragraph{The fourth scale adds cost without a consistent gain.}
At the default step budget, moving from three to four scales increases runtime from $5.46$ to $6.28$\,s, or $15.0\%$. The targeted average improves by only $0.004$ on 2B and decreases by $0.007$ on 8B. This supports using three scales as the default and treating the number of steered scales as an explicit quality--cost control.

\paragraph{Memory on the larger backbone.}
On Infinity-8B, peak memory is $22.7$\,GB with up to three steered scales and rises to $36.1$\,GB at four. Three scales therefore remain the practical default on both model sizes.

\paragraph{Backbone-specific hyperparameters.}
The base step size is $\mathrm{lr}_{\mathrm{base}}=0.005$ on Infinity-2B and $0.045$ on Infinity-8B. The gradient threshold is $\tau_k=10^{-3}$ on 2B and $10^{-12}$ on 8B, reflecting the different gradient scales of the two backbones. At the 8B setting, the threshold is low enough that early stopping and adaptive budget reduction are effectively inactive, so each selected scale uses the full $N_{\max}$. These settings were selected through brief manual exploration on a fixed prompt subset without grid search.

\section{Detailed Comparison with TTS-VAR}
\label{app:ttsvar}

TTS-VAR~\cite{chen2025ttsvar} and VISTA improve frozen next-scale VAR generation through different mechanisms. TTS-VAR searches over multiple sampled trajectories and reallocates computation toward selected prefixes; VISTA instead optimizes the internal representation of one trajectory. This appendix provides the implementation and evaluation details underlying the comparison in Table~\ref{tab:main}.

\paragraph{TTS-VAR mechanism.}
For the 13-scale Infinity-2B model, TTS-VAR uses the descending batch schedule
\[
\{8N,8N,6N,6N,6N,4N,2N,2N,2N,N,N,N,N\}.
\]
At scales 2 and 5, it decodes the current states, extracts DINOv2 features, and uses K-Means++ to retain structurally diverse candidates as the batch contracts. At scales 6 and 9, it decodes the remaining states, scores them with ImageReward, and resamples their prefixes according to the resulting potential scores. TTS-VAR therefore changes which sampled prefixes receive further computation without directly optimizing their representations.

\paragraph{Evaluation protocol.}
For the controlled comparison in Table~\ref{tab:main}, we run the authors' public implementation at $N{=}1$ and $N{=}2$ using the same Infinity-2B backbone, T2I-CompBench prompts, and evaluation code as VISTA. These experiments use one seed. We do not directly substitute the authors' published T2I-CompBench numbers because their reported Infinity baseline and the evaluators used for Non-Spatial and Complex differ from ours. Their $N{=}8$ configuration begins with $64$ coarse candidates and exceeded our available compute budget.

\paragraph{Search and steering address different failure regimes.}
Search is effective when successful generations occur within the sampled candidate set. This is reflected in TTS-VAR's published results: increasing the search budget substantially improves categories such as two-object generation and counting, where outcomes vary across trajectories. Its gains are smaller for positional relations, for which the reported GenEval score changes from $0.2020$ for the backbone to $0.2172$ at $N{=}8$; the authors likewise identify Position as a remaining limitation. Table~\ref{tab:main} shows the same distinction under our evaluation protocol: sampling helps most on categories with variable outcomes, whereas VISTA produces its largest gains on spatial relations by directly modifying the trajectory toward the requested arrangement.

\paragraph{The methods intervene at different stages.}
TTS-VAR reports that reward scores from early intermediate images are weak predictors of final outcomes. It therefore preserves structural diversity through clustering at scales 2 and 5 and applies reward-guided selection only at scales 6 and 9, where score consistency is higher. VISTA requires neither a decoded intermediate image nor a pixel-space verifier: its cross-attention objectives provide a differentiable signal at the coarse scales where scene structure is being established. Search and steering are consequently compatible in principle, but we do not evaluate a combined method.

\paragraph{Resource profiles.}
TTS-VAR's computation and memory grow with its candidate schedule and additionally require intermediate decoding, DINOv2 feature extraction, and ImageReward inference. Its published analysis on an A800-80GB reports peak memory of $30.73$\,GB at $N{=}2$ and $46.40$\,GB at $N{=}4$, compared with $20.34$\,GB for the backbone; the reported timing excludes intermediate image decoding. The paper further reports that BoN with $N{=}8$ could not fit in one pass within $80$\,GB and was therefore executed in two passes.

VISTA maintains one trajectory and uses no auxiliary feature extractor or reward model, but incurs additional forward and backward passes through the backbone. Its cost is controlled by the number of steered scales and optimization steps, as detailed in Appendix~\ref{app:overhead}. Because the published TTS-VAR measurements use a different hardware and timing protocol, we compare the methods' resource scaling rather than claiming a direct wall-clock advantage.

\section{Comparison with Next-Scale Guidance}
\label{app:arguidance}

SSG~\cite{shin2026ssg}, VPG~\cite{liao2026vpg}, and VISTA are training-free inference-time methods evaluated on Infinity-2B, but they optimize different signals. SSG amplifies scale-specific residuals obtained from a frequency-derived coarse prior, targeting refinement and fidelity. VPG contrasts predictions under generated and corrupted prefixes, targeting prefix drift. VISTA instead differentiates prompt-derived compositional objectives through cross-attention.

Neither SSG nor VPG reports T2I-CompBench under the protocol of Table~\ref{tab:main}. Their GenEval results also cannot be inserted as controlled rows in Table~\ref{tab:geneval}: SSG reports only the aggregate score, while VPG uses rewritten prompts and reports values to two decimal places. VISTA uses the original benchmark prompts and averages four seeds. We therefore compare only within-paper changes from each method's own baseline.

\begin{table}[H]
\centering
\caption{Reported Infinity-2B GenEval changes. Each arrow compares a method with its own baseline. Parentheses show absolute changes.}
\label{tab:arguidance}
\small
\setlength{\tabcolsep}{4pt}
\renewcommand{\arraystretch}{1.15}
\begin{tabular}{@{}lcc@{}}
\toprule
Method & Position & Overall \\
\midrule
SSG~\cite{shin2026ssg}
& ---
& \shortstack{$0.683\!\rightarrow\!0.690$\\{\scriptsize $(+0.007)$}} \\

VPG~\cite{liao2026vpg}
& \shortstack{$0.39\!\rightarrow\!0.41$\\{\scriptsize $(+0.02)$}}
& \shortstack{$0.70\!\rightarrow\!0.71$\\{\scriptsize $(+0.01)$}} \\

\rowcolor{vistarow}
VISTA
& \shortstack{$0.250\!\rightarrow\!0.745$\\{\scriptsize $(+0.495)$}}
& \shortstack{$0.672\!\rightarrow\!0.765$\\{\scriptsize $(+0.093)$}} \\
\bottomrule
\end{tabular}
\end{table}

The reported changes reflect the methods' different targets. SSG and VPG modify the generic sampling distribution without encoding a requested relation, whereas VISTA directly optimizes the position constraint measured by GenEval. The larger Position change therefore supports explicit constraint steering, but is not a controlled claim that VISTA is uniformly better in general image quality.